\RequirePackage{fix-cm}
\documentclass{svjour3}
\smartqed
\usepackage{graphicx}

\usepackage{subfigure}
\usepackage{amssymb}
\usepackage{amsmath}
\usepackage{natbib}
\usepackage{algorithm} 
\usepackage{algorithmic}
\usepackage{dsfont}
\usepackage{booktabs}
\usepackage[table]{xcolor}
\usepackage{enumerate}
\RequirePackage[colorlinks=true,linkcolor=green, citecolor=blue,urlcolor=blue]{hyperref}

\newcommand{\crd}[1]{\color{red}#1}
\newcommand{\cbl}[1]{\color{blue}#1}
\newcommand{\cgr}[1]{\color{green}#1}

\newcommand{\argmin}{\arg\min}

\newcommand{\Real}{\mathbb{R}}
\newcommand{\Hcal}{\mathbb{H}}
\newcommand{\X}{\mathcal{X}}
\newcommand{\Y}{\mathcal{Y}}
\newcommand{\Q}{{\boldsymbol{Q}}}
\newcommand{\PM}{{\boldsymbol{P}}}

\newcommand{\I}{{\boldsymbol{I}}}

\newcommand{\K}{{\boldsymbol{K}}}
\newcommand{\D}{{\boldsymbol{D}}}

\newcommand{\x}{\boldsymbol{x}}
\newcommand{\y}{\boldsymbol{y}}
\newcommand{\z}{\boldsymbol{z}}
\newcommand{\vv}{\boldsymbol{v}}
\newcommand{\w}{\boldsymbol{w}}
\newcommand{\e}{\boldsymbol{e}}
\newcommand{\vxi}{\boldsymbol{\xi}}
\newcommand{\valpha}{\boldsymbol{\alpha}}
\newcommand{\vbeta}{\boldsymbol{\beta}}
\newcommand{\vgamma}{\boldsymbol{\gamma}}

\newcommand{\sM}{\mathbb{M}} 
\newcommand{\sB}{\mathbb{B}} 
\newcommand{\sN}{\mathbb{N}} 
\newcommand{\sT}{\mathbb{T}} 
\newcommand{\orcid}[1]{\href{https://orcid.org/#1}{\includegraphics[width=6pt]{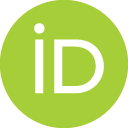}}}

\journalname{``Machine Learning''}
\begin{document}

\title{Unified SVM Algorithm Based on LS-DC Loss
\thanks{This work was supported by the National Natural Science Foundation of China under Grant No. 61772020.}
}

\author{Shuisheng Zhou\orcid{0000-0003-4764-9483}   \and      Wendi Zhou 
}

\authorrunning{S. Zhou, W. Zhou} 

\institute{Shuisheng Zhou \at
              School of Mathematics and Statistics, Xidian University, Xi'An, China 726100 \\
              \email{sszhou@mail.xidian.edu.cn}           
           \and
           Wendi Zhou \at
           School of Computer Science, Beijing University of Posts and Telecommunications, Beijing, China 100867
}

\date{Received: date / Accepted: date}

\maketitle

\begin{abstract}
Over the past two decades, support vector machine (SVM) has become a popular supervised machine learning model, and plenty of distinct algorithms are designed separately based on different KKT conditions of the SVM model for classification/regression with different losses, including convex loss or nonconvex loss.
In this paper, we propose an algorithm that can train different SVM models in a \emph{unified} scheme.
First, we introduce a definition of the \emph{LS-DC} (\textbf{l}east \textbf{s}quares type of \textbf{d}ifference of \textbf{c}onvex) loss and show that the most commonly used losses in the SVM community are LS-DC loss or can be approximated by LS-DC loss.
Based on DCA (difference of convex algorithm), we then propose a unified algorithm, called \emph{UniSVM}, which can solve the SVM model with any convex or nonconvex LS-DC loss, in which only a vector is computed, especially by the specifically chosen loss.
Particularly, for training robust SVM models with nonconvex losses, UniSVM has a dominant advantage over all existing algorithms because it has a closed-form solution per iteration, while the existing algorithms always need to solve an L1SVM/L2SVM per iteration.
Furthermore, by the low-rank approximation of the kernel matrix, UniSVM can solve the large-scale nonlinear problems with efficiency.
To verify the efficacy and feasibility of the proposed algorithm, we perform many experiments on some small artificial problems and some large benchmark tasks with/without outliers for classification and regression for comparison with state-of-the-art algorithms. The experimental results demonstrate that UniSVM can achieve comparable performance in less training time. The foremost advantage of UniSVM is that its core code in Matlab is less than 10 lines; hence, it can be easily grasped by users or researchers.


\end{abstract}

\section{Introduction}\label{sec:intro}

Over the past two decades, support vector machine (SVM) \citep{Vapnik1999, Vapnik2000}, which is based on structural risk minimization, has become a computationally powerful machine learning method for supervised learning. It is widely used in classification and regression tasks \citep{Vapnik2000, Scholkopf2002, Steinwart2008}, such as disease diagnosis, face recognition, and image classification, etc.

Assuming that a training data set $\sT=\{(\x_i, y_i)\}_{i=1}^m$ is drawn independently and identically from a probability distribution on $(\X,\Y)$ with $\X\subset\Real^d$ and $\Y = \{-1, +1\}$ for classification or $\Y = \Real$ for regression, the SVM model aims at solving the following optimization problem:
\begin{equation}\label{eq:svm}
  \w^* \in \argmin_{\w\in\Hcal} \frac\lambda2 \|\w\|^2 + \frac{1}{m}\sum_{i=1}^{m}\ell(y_i, \langle\w,\phi(\x_i)\rangle),
\end{equation}
where $\Hcal$ is a reproducing kernel Hilbert space (RKHS) induced by a kernel function $\kappa(\x,\z) = \langle \phi(\x), \phi(\z)\rangle$ with a feature mapping $\phi:\Real^d\mapsto \Hcal$, $\ell(\cdot,\cdot)$ is a margin-based loss with different choices, and $\lambda$ is the regularizer. The output prediction function $f$ is parameterized by $\w$ as $f(\x) = \langle \w, \phi(\x)\rangle$.
Herein, we take the form without offset for $f$ as in previous papers \citep{Steinwart2003,Keerthi2006, Steinwart2011}. The offset can also be considered by adding an extra attribute 1 to every sample $\x$ or to its feature mapping $\phi(\x)$.

For nonlinear problems, the model \eqref{eq:svm} cannot be solved efficiently because $\phi(\cdot)$ is always a high-dimensional mapping, and even infinite.
By applying the representer theorem \citep{Scholkopf2001,Scholkopf2002,Steinwart2008, Shalev-Shwartz2014}, there exists a vector $\valpha^*\in \Real^m$ such that the solution of \eqref{eq:svm} admits $\w^*=\sum_{i=1}^{m}\alpha^*_i \phi(\x_i).$
Hence, substituting $\w=\sum_{i=1}^{m}\alpha_i \phi(\x_i)$ in \eqref{eq:svm}, we have the following equivalent finite dimensional optimization problem,
\begin{equation}\label{eq:svm_a}
  \min_{\valpha\in\Real^m}  \frac\lambda2 \valpha^\top \K \valpha  + \frac{1}{m}\sum_{i=1}^{m}\ell(y_i, \K_i\valpha),
\end{equation}
where the kernel matrix $\K$ satisfies $\K_{i,j}=\kappa(\x_i,\x_j)$ and $\K_i$ is the $k$-th row of $\K$.
{The similar model of \eqref{eq:svm_a} can also be derived by duality \citep{Vapnik2000,BoydV2009}, where the coefficients $\valpha$ may be properly bounded (see \eqref{eq:l1svm}, \eqref{eq:l2svm} and \eqref{eq:svr} for details).}

Many scholars have studied different SVM models based on different loss functions.
The typical works \citep{Vapnik2000,Suykens1999,Keerthi2006,Zhou2009,Steinwart2011,Zhou2013, Zhou2013a,Zhou2016} are focused on SVM models with convex loss, such as L1-SVM with the hinge loss, L2-SVM with the squared hinge loss, LS-SVM with the least squares least loss, and the support vector regression (SVR) with $\varepsilon$-insensitive loss, etc.
The state-of-the-art SVM tool, \texttt{LibSVM} (including SVC and SVR) \citep{Chang2011}, covers some cases with convex losses and has numerous applications.

The algorithms based on convex losses are sensitive to outliers, where ``outlier'' refers to the contaminated samples far away from the majority instances with the same labels \citep{Hampel2011}, which may emerge through mislabeling. This is because the contaminated data have the largest weights (support values) to represent the output function in this case.

Many researchers have used nonconvex loss functions to weaken the influence of outliers. For example, \citet{Shen2003}, \citet{Collobert2006}, and \citet{Wu_Liu2007} study the robust SVM with the truncated hinge loss; \citet{Tao2018} study the robust SVM with the truncated hinge loss and the truncated squared hinge loss. 
Based on DCA (difference of convex algorithm) procedure \citep{Thi2018, Yuille2003}, all those studies have presented algorithms to \emph{iteratively} solve L1SVM/L2SVM to obtain the solutions of their proposed nonconvex models.
By introducing the smooth nonconvex losses, \citet{Feng2016} propose the robust SVM models which solve a re-weighted L2SVM many times. See subsection \ref{subsec:review_SVM2} for the representative robust SVMs.

All the robust SVM algorithms mentioned above have double-layer loops. The inner loop is used to solve a convex problem with parameters adjustable by the outer loop, and the outer loop adjusts those parameters to reach the solution of the nonconvex model.

However, the inner loop of these algorithms is computationally expensive. For example, \citet{Collobert2006, Wu_Liu2007, Feng2016, Tao2018} solve a constrained quadratic programming (QP) defined by L1SVM, L2SVM or re-weighted L2SVM, and all state-of-the-art methods for quadratic programming require substantial numbers of iterations {(Such as SMO \citep{Platt1999,Keerthi2001,Chen2006} or the tools \texttt{quadprog} in Matlab)}. In \citet{Tao2018}, some efficient techniques based on the coordinate descent are given to reduce the cost of the inner loop, but it remains necessary to solve L1SVM/L2SVM, perhaps with smaller size.

There are three weaknesses to the existing algorithms of the robust SVM models. The first is that the total computational complexity is high, resulting in long training time which limits the algorithms in processing large-scale problems.
The second is that most of the existing algorithms are only suitable for classification problems and require complicated modifications when applied for regression problems.
The third is that all the existing algorithms are designed separately based on the special kinds of losses, thus costing much effort for the readers/users to learn the different algorithms or to change the losses before making use of them.

Recently, \citet{Chen2018} proposed the robust LSSVM based on the truncated least squares loss, which partly resolves the first two weaknesses (without inner loop and solving classification/regression task similarly).
To extend this benefit to all the other losses, by defining an \textbf{LS-DC loss}, we propose a unified solution for different models with different losses, named as \textbf{UniSVM}, which overcomes all three mentioned weaknesses.

Here, we only focus on the positive kernel case, namely, where $\Hcal$ is an RKHS. For the nonpositive kernel case, \citet{Ong2004} generalized this type of learning problem to reproducing kernel Kre\u{i}n spaces (RKKS) and verified that the representer theorem still holds in RKKS even if its regularization part is nonconvex. Recently, \citet{Xu2017} decomposed the regularization part as DC form and proposed an efficient learning algorithm, where the loss is chosen as only the squared hinge loss. Our results in this work can be seamlessly generalized to nonpositive case by the methods in \citet{Xu2017}.

Our contributions in this work can be summarized as follows:
\begin{itemize}
\item We define a kind of loss with a DC decomposition, called \textbf{LS-DC loss}, and show that all the commonly used losses are LS-DC loss or can be approximated by LS-DC loss.
\item We propose a UniSVM algorithm which can deal with any LS-DC loss in a unified scheme, including convex or nonconvex loss, classification or regression loss, in which only one vector is dominated by the specifically chosen loss. Hence, it can train the classification problems and the regression problems in the same scheme.
\item The proposed UniSVM has low computational complexity, even for nonconvex models, because it solves a system of linear equations iteratively, which has a closed-form solution. Hence, the inner loop disappears.
\item By the efficient low-rank approximation of the kernel matrix, UniSVM can solve the large-scale problem efficiently.
\item In view of the theories of DCA, UniSVM converges to the global optimal solution of the convex model, or to a critical point of the nonconvex model.
\item UniSVM can be easily grasped by users or researchers because its core code in Matlab is less than 10 lines.
\end{itemize}

The notations in this paper are as follows. All the vectors and matrices are in bold style, such as $\vv,\x_i$ or $\K$, and the set or space is noted as $\sM, \sB$, $\Real^m$, etc. The scalar $v_i$ is the $i$-th element of $\vv$, the row vector $\K_i$ is the $i$-th row of $\K$, and $\K_\sB$ is the submatrix of $\K$ with all \emph{rows} in the index set $\sB$. The {transpose} of the vector $\vv$ or matrix $\K$ is noted as $\vv^\top$ or $\K^\top$. $\I$ is an identity matrix with proper dimensions,  $t_+:=\max\{t,0\}$, and $\mathds{1}_{a} =1$ if the event $a$ is true, and otherwise 0.

The rest of the paper is organized as follows.
In Section \ref{sec:rev}, we review the DCA procedure and the related SVM models.
In Section \ref{sec:lsdc_loss}, we define an LS-DC loss which has a desirable DC decomposition and reveal its properties.
In Section \ref{sec:UniSVM}, we propose a UniSVM algorithm that can train the SVM model with any different LS-DC loss based on DCA.
In Section \ref{sec:exp}, we verify the effectiveness of UniSVM by many experiments, and Section \ref{sec:con} concludes the paper.

\section{Reviews of related works}\label{sec:rev}

We review the DCA procedure and some SVM models with convex and nonconvex loss in this section.

\subsection{DC programming and DCA}\label{subsec:review_DCA}

As an efficient nonconvex optimization technique, DCA, first introduced in \citet{Tao1986} and recently reviewed in \citet{Thi2018}, has been successfully applied in machine learning \citep{Yuille2003, Neumann2004,Collobert2006,Le2008,Le2009,Ong2013,Xu2017, Tao2018,Chen2018}. A function $F(x)$ is called a difference of convex (DC) function if $F(\x)=H(\x)-G(\x)$, with $H(\x)$ and $G(\x)$ being convex functions.
Basically, DC programming is used to solve
\begin{equation}\label{eq:dc}
  \min_{\x\in\X} F(\x):= H(\x)-G(\x),
\end{equation}
where $H(\x)$ and $G(\x)$ are convex functions and $\X$ is a convex set.

DCA is a majorization-minimization algorithm \citep{Naderi2019} which works by optimizing a sequence of upper-bounded convex functions of $F(\x)$. For the current approximated solution $\x^k$, since $G(\x)\geq G(\x^k)+\langle \x-\x^k, \vv^k\rangle$ with $\vv^k\in\partial G(\x^k)$, $H(\x)-\langle\vv^k,\x-\x^k\rangle - G(\x^k)$ is an upper-bounded convex function of $F(\x)$. Thus, to solve the DC problem \eqref{eq:dc}, DCA iteratively obtains a new solution $\x^{k+1}$ by {solving the convex programming as follows.}
\begin{equation}\label{eq:dca}
  \x^{k+1}\in\arg\min_{\x} H(\x)-\langle\vv^k,\x\rangle.
\end{equation}
There is a convergence guarantee \citep{Thi2018}. {Particularly, if $H(\x)$ is a quadratic function, the optimal problem \eqref{eq:dca} has a closed form solution. Thus, the DCA procedure has no inner iterations. This motivates us to design a DC decomposition for the losses of SVMs models to speed up the algorithm in Section \ref{sec:lsdc_loss}. }

\subsection{SVM models with convex losses and nonconvex losses}
\subsubsection{SVM models with convex losses}\label{subsec:review_SVM1}
If the hinge loss $\ell(y,t):=\max\{0,1-yt\}$ is chosen in \eqref{eq:svm} for classification, then L1SVM is obtained by duality as following \citep{Vapnik2000,Vapnik1999, Keerthi2006, Steinwart2011, Zhou2013}:
\begin{equation}\label{eq:l1svm}
{\rm L1SVM:~}\min_{0\leq\vbeta\leq \frac1m} \frac1{2\lambda}\vbeta^\top\widetilde\K\vbeta -\e^\top \vbeta,
\end{equation}
where $\widetilde\K_{i,j} = y_iy_j\K_{i,j}$ and $\e=(1,\cdots, 1)^\top\in\Real^m$. If the squared hinge loss $\ell(y,t):=\frac12\max\{0,1-yt\}^2$ is chosen in \eqref{eq:svm}, then L2SVM is obtained by duality as following \citep{Vapnik2000,Vapnik1999, Steinwart2011, Zhou2013, Zhou2013a,Zhou2009}:
\begin{equation}\label{eq:l2svm}
{\rm L2SVM:~} \min_{0\leq\vbeta} \frac1{2\lambda}\vbeta^\top\left(\widetilde\K+\lambda m\I\right)\vbeta -\e^\top \vbeta,
\end{equation}
where $\I$ is the identity matrix. With the solution $\vbeta^*$, the unknown sample $\x$ is predicted as $sgn(f(\x))$ with $f(\x) = \frac1\lambda\sum_{i=1}^{m}y_i\beta_i^*\kappa(x,x_i)$ for model \eqref{eq:l1svm} or \eqref{eq:l2svm}.

If the least squares loss $\ell(y,t):=\frac12(1-yt)^2 = \frac12(y-t)^2$ is chosen in \eqref{eq:svm}, the LSSVM model obtained by duality \citep{Suykens1999,Suykens1999a,Suykens2002,Jiao2007} is
\begin{equation}\label{eq:lssvm}
{\rm LSSVM:~} \min_{\vbeta} \frac1{2\lambda}\vbeta^\top\left(\K+\lambda m\I\right)\vbeta -\y^\top \vbeta,
\end{equation}
with a unique nonsparse solution, where $\y=(y_1,\cdots,\y_m)^\top$. Recently, also by choosing the least squares loss in \eqref{eq:svm}, \citet{Zhou2016} proposed the primal LSSVM (PLSSVM) based on the representer theorem\footnote{For consistency with the model induced by duality, we let $\w=\frac1\lambda\sum_{i=1}^{m}\vbeta_i \phi(\x_i)$ here.} as
\begin{equation}\label{eq:plssvm}
{\rm PLSSVM:~} \min_{\vbeta} \frac1{2\lambda}\vbeta^\top\left(m\lambda\K+\K\K^\top\right)\vbeta -\y^\top\K \vbeta,
\end{equation}
which may have a sparse solution if $\K$ has low rank or can be approximated as a low rank matrix.
With the solution $\vbeta^*$, the unknown sample $\x$ is predicted as $sgn(f(\x))$(classification) or $f(\x)$ (regression) with $f(\x) = \frac1\lambda\sum_{i=1}^{m}\beta_i^*\kappa(x,x_i)$ for model \eqref{eq:lssvm} or \eqref{eq:plssvm}.

If the $\varepsilon$-insensitive loss $\ell_\varepsilon(y,t):= (|y-t|-\varepsilon)_+$ is chosen in \eqref{eq:svm} for the regression problem, then SVR is obtained as
\begin{equation}\label{eq:svr}
{\rm SVR:~} \min_{0\leq\vbeta,\hat\vbeta\leq \frac1m} \frac1{2\lambda}(\vbeta-\hat\vbeta)^\top \K (\vbeta-\hat\vbeta) +\varepsilon\sum_{i=1}^{m}(\vbeta+\hat\vbeta) + \sum_{i=1}^{m}y_i(\vbeta-\hat\vbeta),
\end{equation}
and with the solution $(\vbeta^*,\hat\vbeta^*)$, the prediction of the new input $\x$ is  $f(\x) = \frac1\lambda\sum_{i=1}^{m}(\beta_i^*-\hat\beta_i^*)\kappa(x,x_i).$

\subsubsection{Robust SVM models with nonconvex losses}\label{subsec:review_SVM2}
To improve the robustness of the L1SVM model \eqref{eq:l1svm}, in \citet{Collobert2006} the hinge loss $(1-yt)_+$ in \eqref{eq:svm} is truncated as the ramp loss $\min\{(1-yt)_+, a\}$ with $a>0$ and decomposed it as DC form $(1-yt)_+ - (1-yt-a)_+$. {The problem \eqref{eq:svm} is posed as a DC programming:
$$\min_{\w\in\Hcal} \frac\lambda2 \|\w\|^2 + \frac{1}{m}\sum_{i=1}^{m}(1-y_i\langle\w,\phi(\x_i)\rangle)_+ - \frac{1}{m}\sum_{i=1}^{m}(1-y_i\langle\w,\phi(\x_i)\rangle - a)_+.$$
Then, based on DC procedure and Lagrange duality, a robust L1SVM is proposed by iteratively solving the following L1SVM problem at the $(k+1)-$th iteration.
\begin{equation}\label{eq:rl1svm}
\vbeta^{k+1}\in\arg\min_{0\leq\vbeta\leq\tfrac1m} \frac1{2\lambda}(\vbeta-\vv^{k})^\top\widetilde\K(\vbeta-\vv^{k}) -\e^\top \vbeta,
\end{equation}
where $\vv^{k}$ satisfies $v^{k}_i = \frac1m\cdot\mathds{1}_{1-\widetilde\K_i(\vbeta^{k}-\vv^k)/\lambda>a}, i=1,\cdots,m$. Similar analysis with a different form appears in \citet{Wu_Liu2007} and \citet{Tao2018}.

To improve the robustness of L2SVM \eqref{eq:l2svm}, \citet{Tao2018} truncated the squared hinge loss $(1-yt)_+^2$ as $ \min\{(1-yt)_+^2, a\}$ in \eqref{eq:svm} and decomposed it as DC form $(1-yt)_+^2-\left( (1-yt)_+^2 - a\right)_+$.
Based on DCA, the robust solution of L2SVM is given by iteratively solving
\begin{equation}\label{eq:rl2svm}
\vbeta^{k+1}\in\arg\min_{\vbeta\geq0} \frac1{2\lambda}(\vbeta-\vv^{k})^\top\widetilde\K(\vbeta-\vv^{k}) +\frac m{2}\vbeta^\top\vbeta -\e^\top \vbeta,
\end{equation}
where $\vv^{k}$ satisfies $v^{k}_i = \frac1m(1-\tfrac1\lambda\widetilde\K_i(\vbeta^{k}-\vv^k))\cdot\mathds{1}_{1-\widetilde\K_i(\vbeta^{k}-\vv^k)/\lambda>\sqrt{a}}, i=1,\cdots,m$.
However, the analysis in \citet{Tao2018} pointed out that \eqref{eq:rl2svm} is not a ``satisfactory learning algorithm'' in this case. By deleting the current outliers, which satisfy $1-y_i f(\x_i)>\sqrt a$ from the training set iteratively, \citet{Tao2018} proposed a multistage SVM (MS-SVM) to solve the robust L2SVM. Concisely, they first solve an original L2SVM \eqref{eq:l2svm}, and then some smaller L2SVMs iteratively.

Although \citet{Tao2018} propose some improved methods (coordinate descent and an inexpensive scheme), all of the given algorithms still solve a constrained QP per iteration, and possibly with smaller size.

In contrast, \citet{Feng2016} propose a robust SVM model, in which the following smooth nonconvex loss
\begin{eqnarray}\label{eq:los2}
  \ell_a(y,t) = a\left(1-\exp\left(-\tfrac1a(1-yt)_+^2\right)\right)
\end{eqnarray}
{with $a> 0$ is chosen in \eqref{eq:svm}. The loss \eqref{eq:los2} is approximated as the squared hinge loss $(1-yt)_+^2$ when $a\rightarrow+\infty$ and can be considered as a smooth approximation of the truncated squared hinge loss $ \min\{(1-yt)_+^2, a\}$.}
After analysis of the KKT conditions of the given model, \citet{Feng2016}
put forward the algorithm by solving the following re-weighted L2SVM iteratively
{\begin{equation}\label{eq:wl2svm}
  \vbeta^{k+1} \in \argmin_{\vbeta\geq 0}  \tfrac1{2\lambda}\vbeta^\top\left(\widetilde\K+m\lambda\D^k\right)\vbeta -\e^\top \vbeta
\end{equation}
where $\D^k$ is a diagonal matrix satisfying $\D_{i,i}^k=\left(\psi'((1-\widetilde\K_i\vbeta^k)_+^2)\right)^{-1}$ with $\psi(u)= a\left(1-\exp(-\tfrac1 au)\right)$.
}

All of those algorithms for robust SVM models \citep{Collobert2006,Wu_Liu2007,Tao2018,Feng2016} must solve a constrained QP in the inner loops, which results in the long training time.

{Based on the decomposition of the truncated \emph{least squares loss} $\min\{(1-yt)^2,a\}$ in \eqref{eq:svm} or \eqref{eq:svm_a} as $(1-yt)^2-((1-yt)^2-a)_+$ and the representer theorem, the robust sparse LSSVM (RSLSSVM) was studied in \citet{Chen2018} by solving
\begin{equation}\label{eq:sqsvmdc}
  \valpha^{k+1} \in \argmin_{\valpha\in\Real^m} \frac\lambda2\valpha^\top\K\valpha +\tfrac{1}{2m}\sum_{i=1}^{m} (y_i -\K_i\valpha)^2 - \tfrac1{m}\langle\K\vv^k, \valpha\rangle,
\end{equation}
with $\vv^k$ as $v^k_i = -(y_i-\K_i\valpha^k)\mathds{1}_{|1-y_i\K_i\valpha^k|>\sqrt a}$.

The model \eqref{eq:sqsvmdc} for solving the nonconvex SVM has three advantages. The first is that it has a closed-form solution since it is an  unconstrained QP. Second, it has a sparse solution if $\K$ has low rank or can be approximated by a low rank matrix (see \citep{Zhou2016} for details). Thus, it can be solved with efficiency. Furthermore, \eqref{eq:sqsvmdc} can also be applied to regression problems directly.}

To extend those benefits to all the other losses for the classification task and regression task simultaneously, we define LS-DC loss in Section \ref{sec:lsdc_loss} and propose a unified algorithm in Section \ref{sec:UniSVM}, which includes all SVM models (including the classification/regression SVM with convex loss and nonconvex loss) in a unified scheme.

\section{LS-DC loss function}\label{sec:lsdc_loss}
Here, we first define a kind of loss called LS-DC loss, and then show that most popular losses are in fact LS-DC loss or can be approximated by LS-DC loss.

For any margin-based loss $\ell(y,t)$ of SVM, let $\psi(u)$ satisfy $\psi(1-yt):=\ell(y,t)$ for classification loss or $\psi(y-t):=\ell(y,t)$ for regression loss. To obtain a useful DC decomposition of the loss $\ell(y,t)$, we propose the following definition.
\begin{definition}[LS-DC loss]\label{def:lsdc_loss}
\emph{We call $\ell(y,t)$ \textbf{a least squares type DC loss}, abbreviated as \textbf{LS-DC loss}, if there exists constant $A$ ($0<A<+\infty$) such that $\psi(u)$ has the following DC decomposition:
\begin{equation}\label{eq:lsdc_loss}
  \psi(u) = Au^2 - (Au^2-\psi(u)).
\end{equation}}
\end{definition}
The essence of the definition demands $Au^2-\psi(u)$ to be convex. The following theorem is clear.
\begin{theorem}\label{th:th1}
  If the loss $\psi(u)$ is second-order derivable and $\psi''(u)\leq M$, then it is an LS-DC loss with parameter $A\geq \frac M 2$.
\end{theorem}

Not all losses are LS-DC losses, even the convex losses. We will show that the hinge loss and the $\varepsilon$-insensitive loss are not LS-DC losses. However, they can be approximated by LS-DC losses.

Next, we will show that most losses used in the SVM community are LS-DC losses or can be approximated by LS-DC loss. The proofs are listed in Appendix \ref{appendix:proof}.

\begin{proposition}[LS-DC property of classification losses]\label{subsec:class_loss}
  The most commonly used classification losses are cases of LS-DC loss or can be approximated by LS-DC losses. We enumerate them as follows.
  \begin{enumerate}[(a)]
  \item The least squares loss $\ell(y,t) = (1-yt)^2$ is an LS-DC loss {with $Au^2-\psi(u)$ vanished}. \label{item:1}
  \item The truncated least squares loss $\ell(y,t) =\min\{(1-yt)^2,a\}$ is an LS-DC loss with $A\geq1$. \label{item:2}
  \item The squared hinge loss $\ell(y,t)=(1-yt)_+^2$ is an LS-DC loss with $A \geq 1$.\label{item:4}
  \item The truncated squared hinge loss $\ell(y,t) =\min\{(1-yt)_+^2,a\}$
       is an LS-DC loss with $A\geq 1$.  \label{item:5}
  \item The hinge loss $\ell(y,t) = (1-yt)_+$ is \textbf{not} an LS-DC loss. However, if it is approximated as $\frac1p\log(1+\exp(p(1-yt)))$ with a finite $p$, we obtain an LS-DC loss with $A\geq p/8$. \label{item:6}
  \item The ramp loss $\ell(y,t) =\min\{(1-yt)_+, a\}$ is also \textbf{not} an LS-DC loss. However, we can give two smoothed approximations of the ramp loss:
      \begin{eqnarray}
         \ell_{a}(y,t) &=&\left\{\begin{array}{ll}
                            \frac2a (1-yt)_+^2, & 1-yt \leq \frac a2, \\
                            a-\frac2a (a-(1-yt))_+^2, & 1-yt > \frac a2.
                          \end{array}
         \right. \label{eq:ramp_s} \\
          \ell_{(a,p)}(y,t) &=&
           \frac1p\log\left(\frac{1+\exp(p(1-yt))}{1+\exp(p(1-yt-a))}\right). \label{eq:ramp_sp}
    \end{eqnarray}
    The first one has the same support set as the ramp loss, and the second one is derivable with any order. The loss \eqref{eq:ramp_s} is an LS-DC loss with $A\geq 2/a$ and \eqref{eq:ramp_sp} is an LS-DC loss with $A\geq p/8$.

  \item The nonconvex smooth loss \eqref{eq:los2} proposed in \cite{Feng2016} is an LS-DC loss with $A\geq 1$.
   \item Following the nonconvex smooth loss \eqref{eq:los2}, we generalize it as \label{item:nonconvex_loss}
       \begin{eqnarray}\label{eq:los4}
         \ell_{(a,b,c)}(y,t) = a\left(1-\exp\left(-\tfrac1b(1-yt)_+^c\right)\right),
       \end{eqnarray}
      where $a,b>0, c\geq2$. The loss \eqref{eq:los4} is an LS-DC loss with the parameter $A\geq\frac12{M(a,b,c)}$, where
      \begin{equation}\label{eq:M_abc}
        M(a,b,c):= \tfrac{ac}{b^{2/c}}\left((c-1)(h(c))^{1-2/c}-c (h(c))^{2-2/c}\right)e^{-h(c)},
      \end{equation}
      with $h(c):=\left(3(c-1)-\sqrt{5c^2-6c+1}\right)/(2c)$.
\end{enumerate}
\end{proposition}

Regarding the new proposed loss \eqref{eq:los4}, we offer the following comments.

\begin{remark}
Two more parameters are introduced to the loss \eqref{eq:los4} to make it more flexible; the parameter $a$, which is the limitation of the loss function if $1-yt\rightarrow \infty$, describes the effective value (or saturated value) of the loss function for large inputs; the parameter $b$, which characterizes the localization property of the loss functions, describes the rate of the loss function saturated to its maximum and minimum. By uncoupling $a$ and $b$, we improve the flexibility of the robust loss. For example, the inflection point of \eqref{eq:los2} is $yt=1-\sqrt{a/2}$, which is directly controlled by the saturated value $a$, and the inflection point of \eqref{eq:los4} is $yt=1-\sqrt{b/2}$ if $c=2$, which is only controlled by the parameter $b$.
In experiments, by simply adjusting $a$, $b$, and $c$, we can obtain better performance.
\end{remark}

The most commonly used losses for classification are summarized in Table \ref{tab:classification_loss} in Appendix \ref{appendix_a}. Some classification losses and their LS-DC decompositions are also plotted in Figure \ref{fig_lossC}.

\begin{figure}[htbp]
\centering
\subfigure[Truncated least squares loss $\min\{u^2,1\}$.]{
\includegraphics[width=0.31\textwidth]{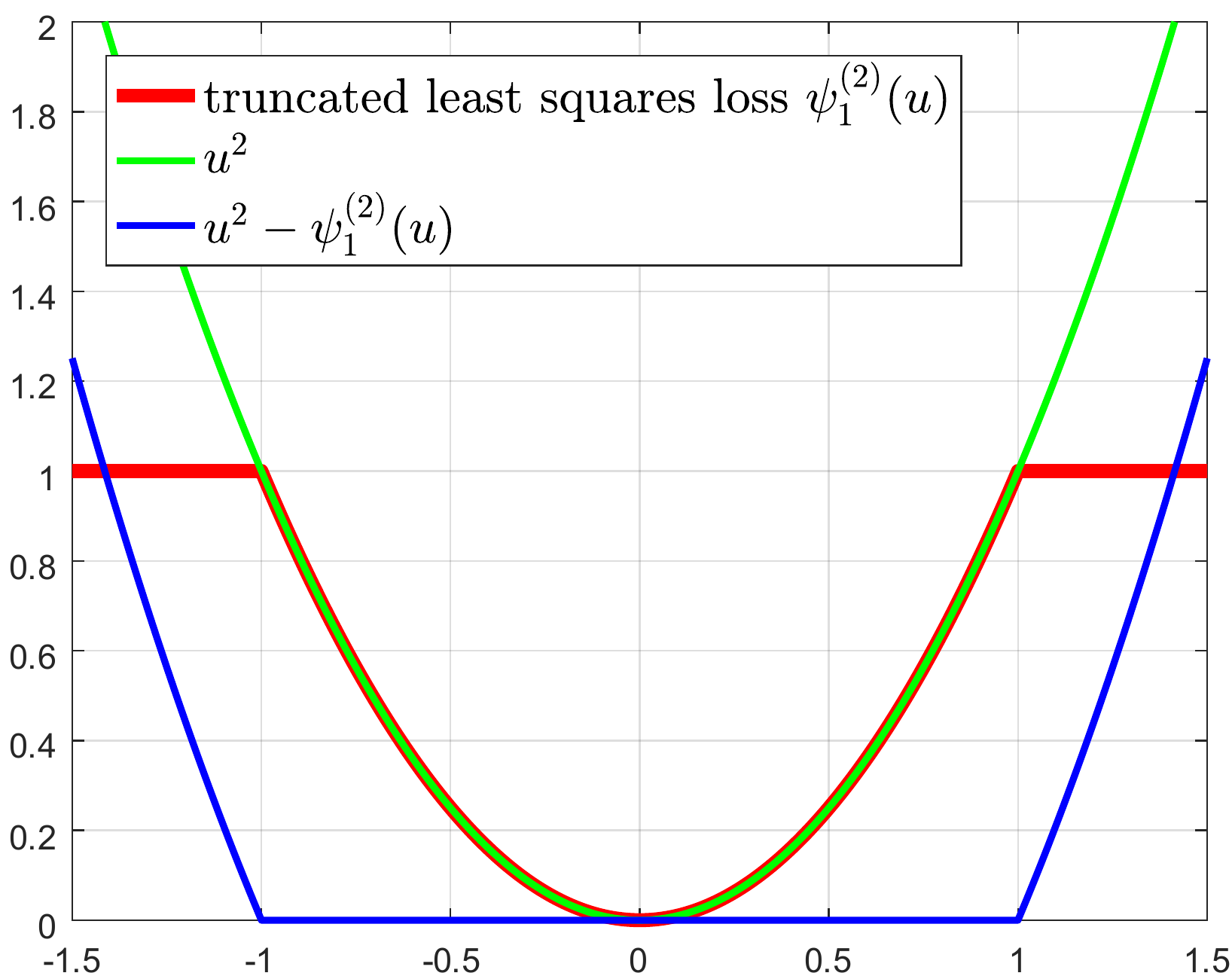}
}
\subfigure[Squared hinge $u^2_+$.]{
\includegraphics[width=0.31\textwidth]{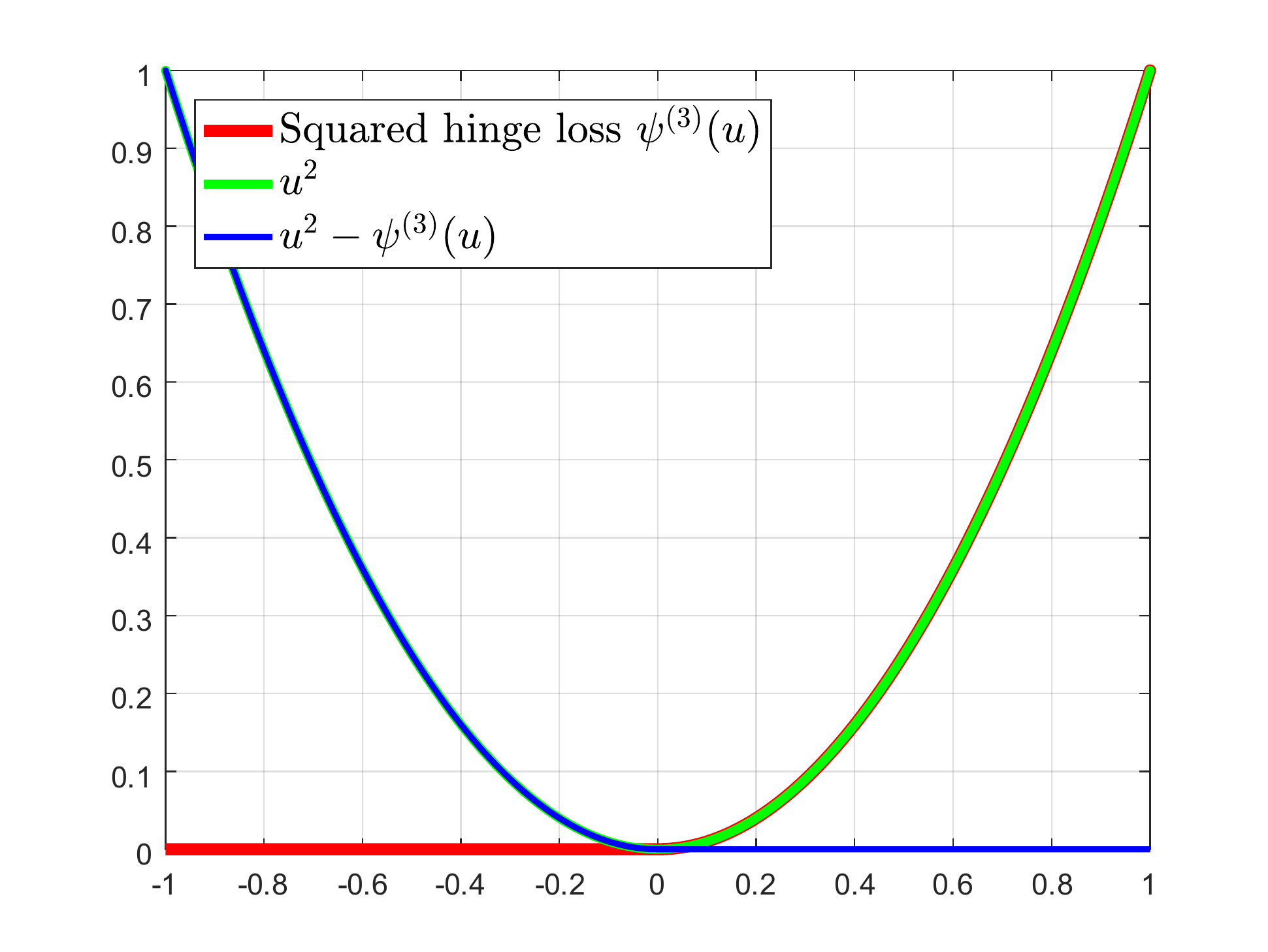}
}
\subfigure[Truncated squared hinge loss $\min\{u^2_1, 1\}$.]{
\includegraphics[width=0.31\textwidth]{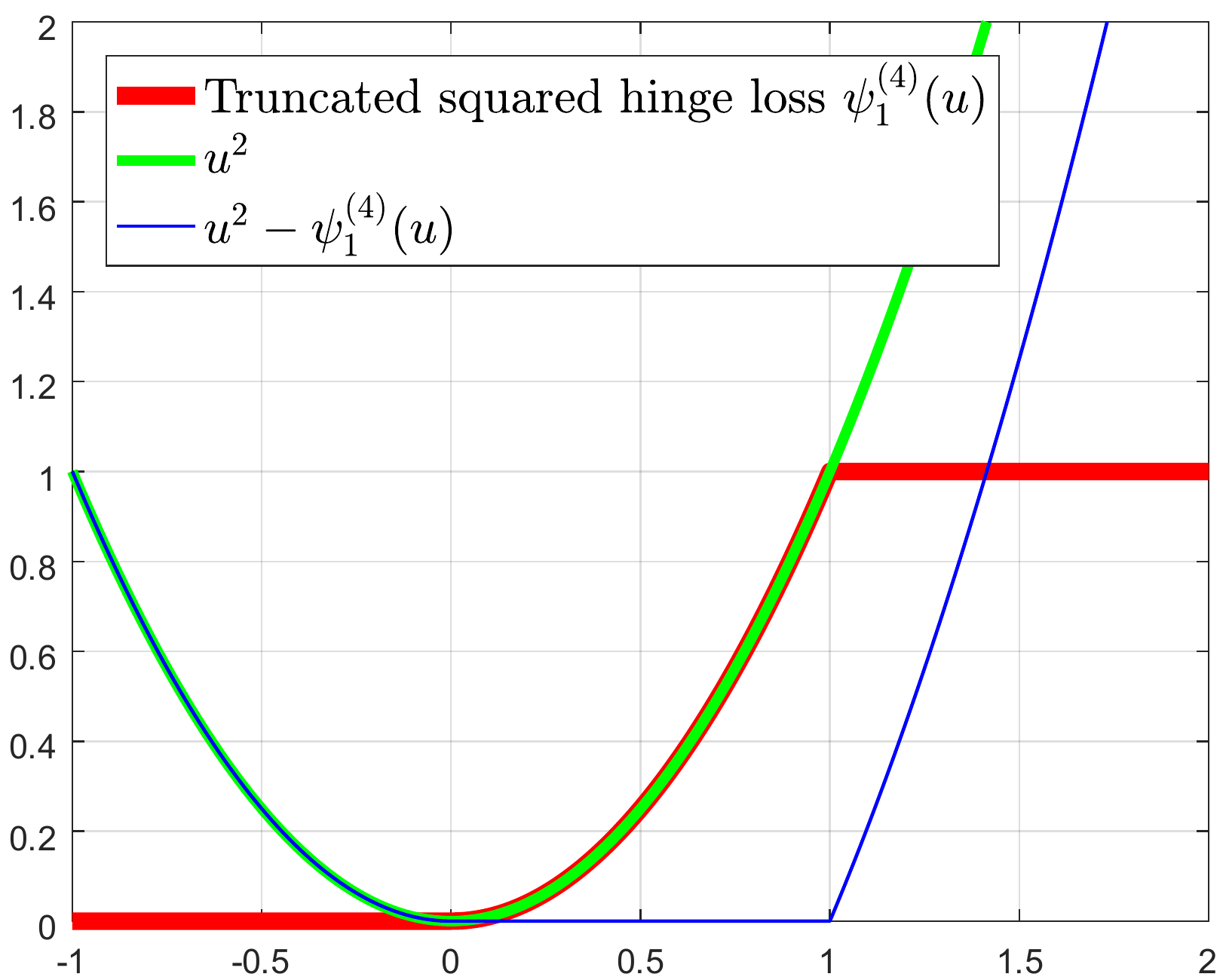}
}

\subfigure[Hinge loss and its approximation $\frac18\log(1+\exp(8u))$.]{
\includegraphics[width=0.31\textwidth]{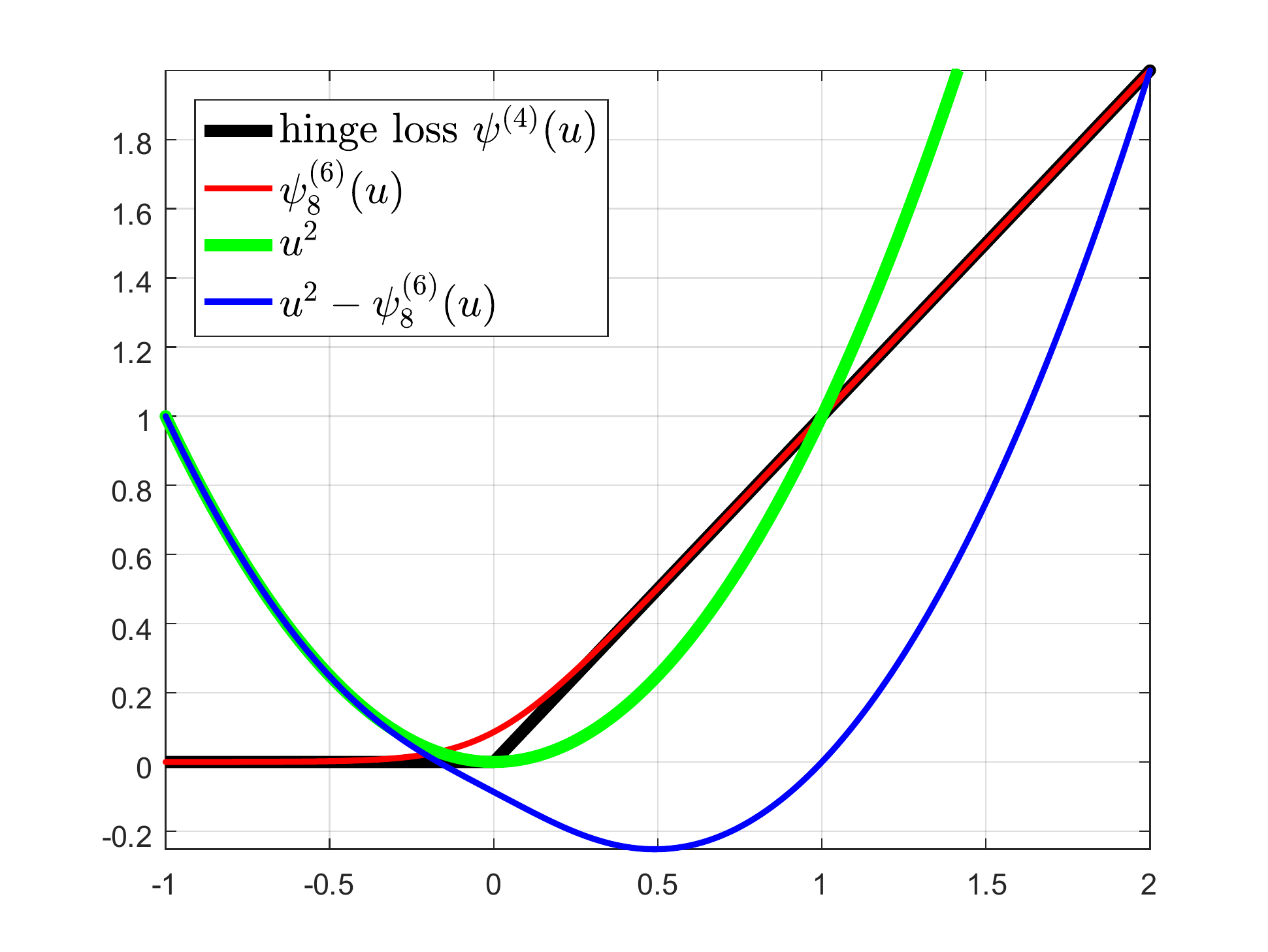}
}
\subfigure[Smoothed ramp loss \eqref{eq:ramp_s} with $a=1$.]{
\includegraphics[width=0.31\textwidth]{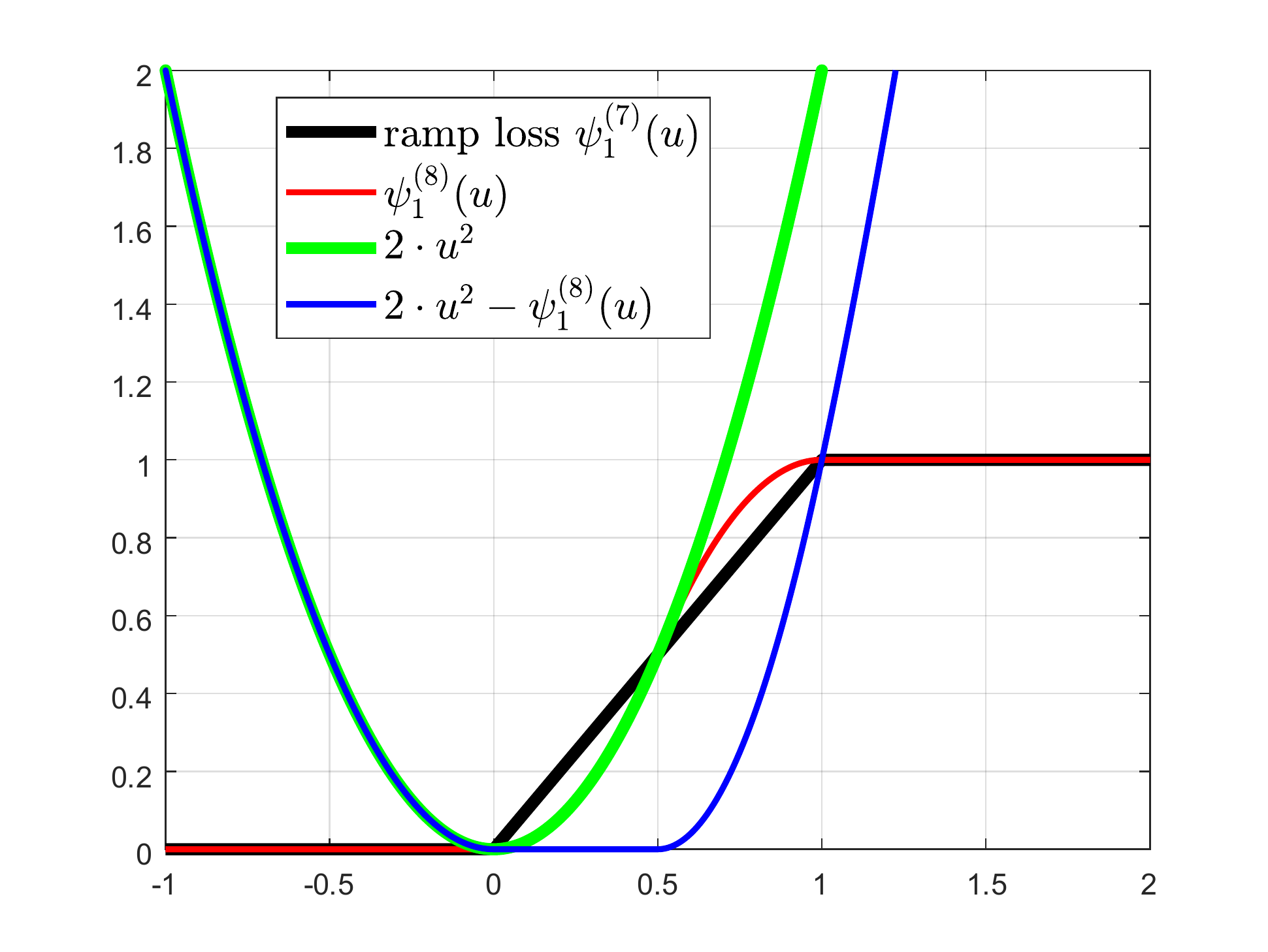}
}
\subfigure[Smoothed ramp loss \eqref{eq:ramp_sp} with $a=1, p=8$.]{
\includegraphics[width=0.31\textwidth]{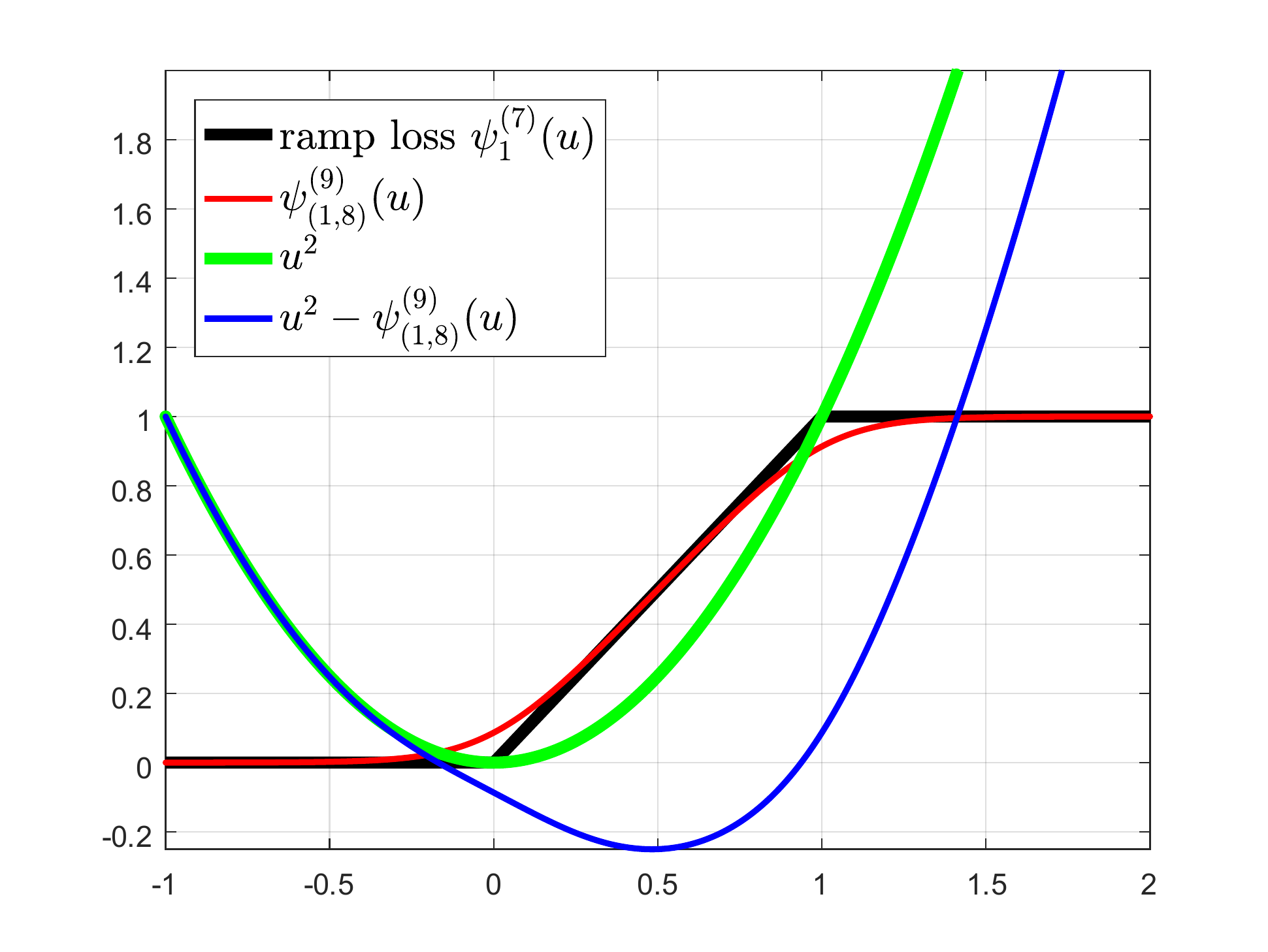}
}

\subfigure[Smoothed nonconvex loss \eqref{eq:los4} with $a=b=1,c=2$.]{
\includegraphics[width=0.31\textwidth]{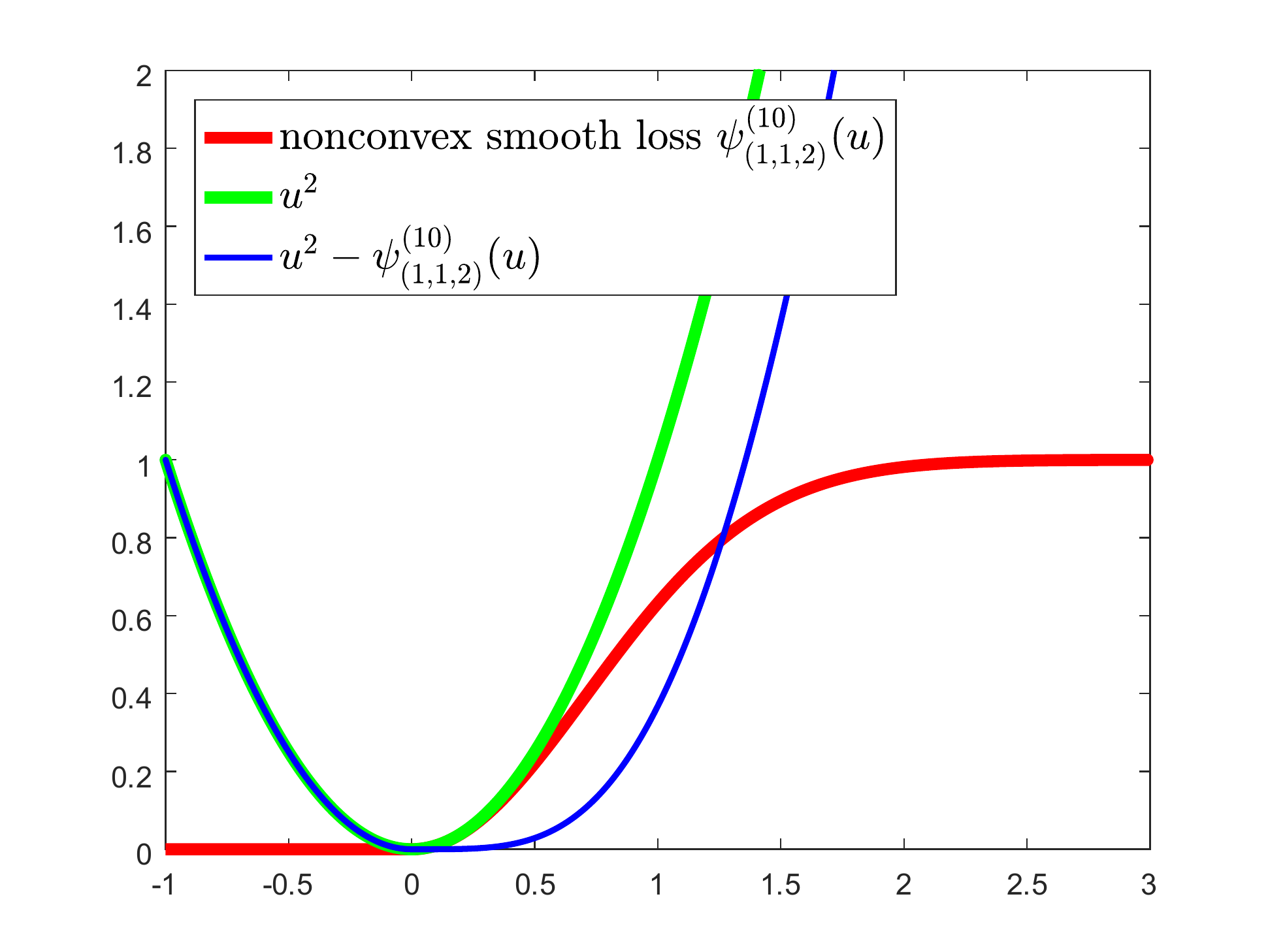}
}
\subfigure[Smoothed nonconvex loss \eqref{eq:los4} with $a=1,b=c=2$.]{
\includegraphics[width=0.31\textwidth]{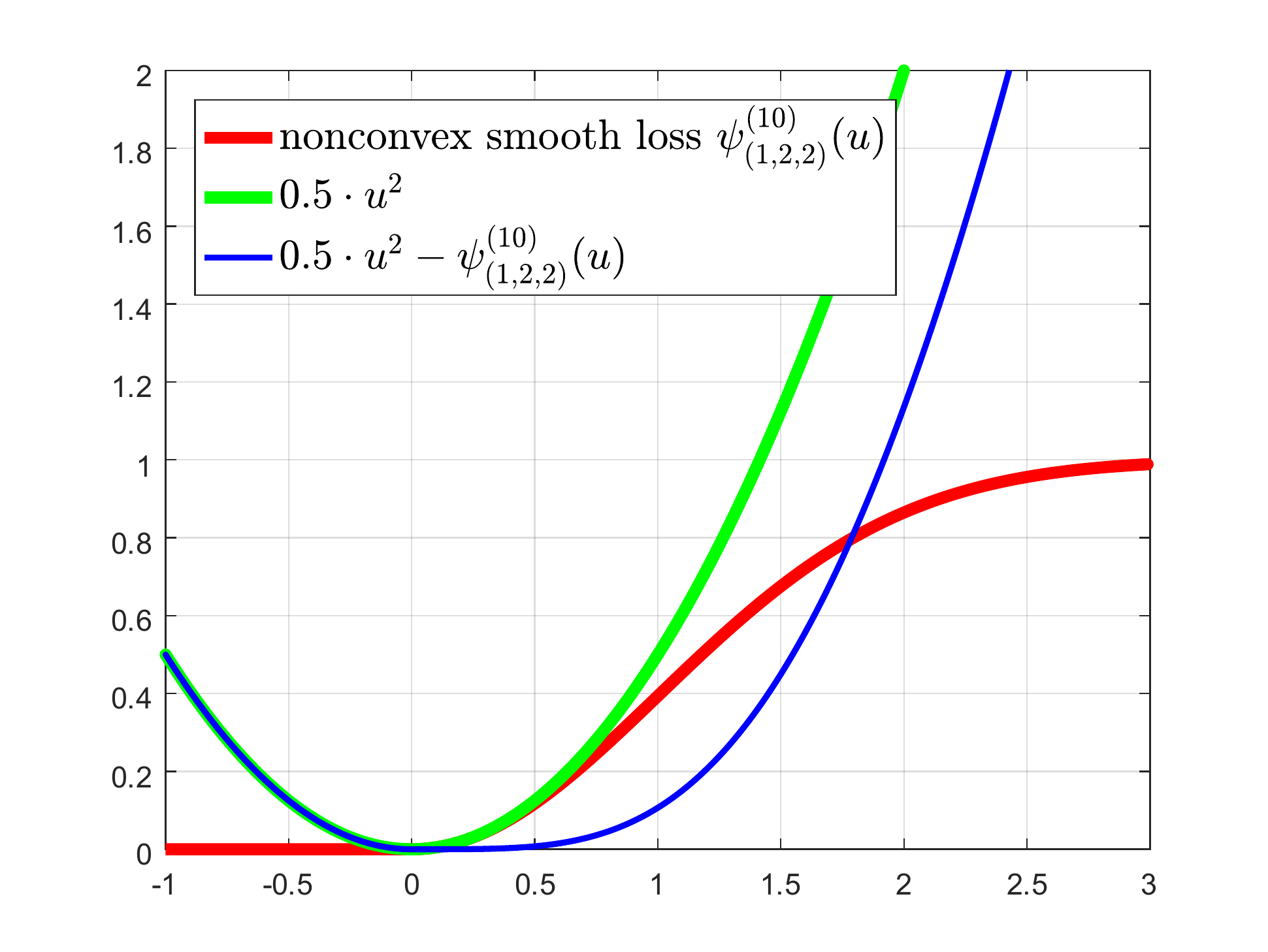}
}
\subfigure[Smoothed nonconvex loss \eqref{eq:los4} with $a=1,b=2,c=4$]{
\includegraphics[width=0.31\textwidth]{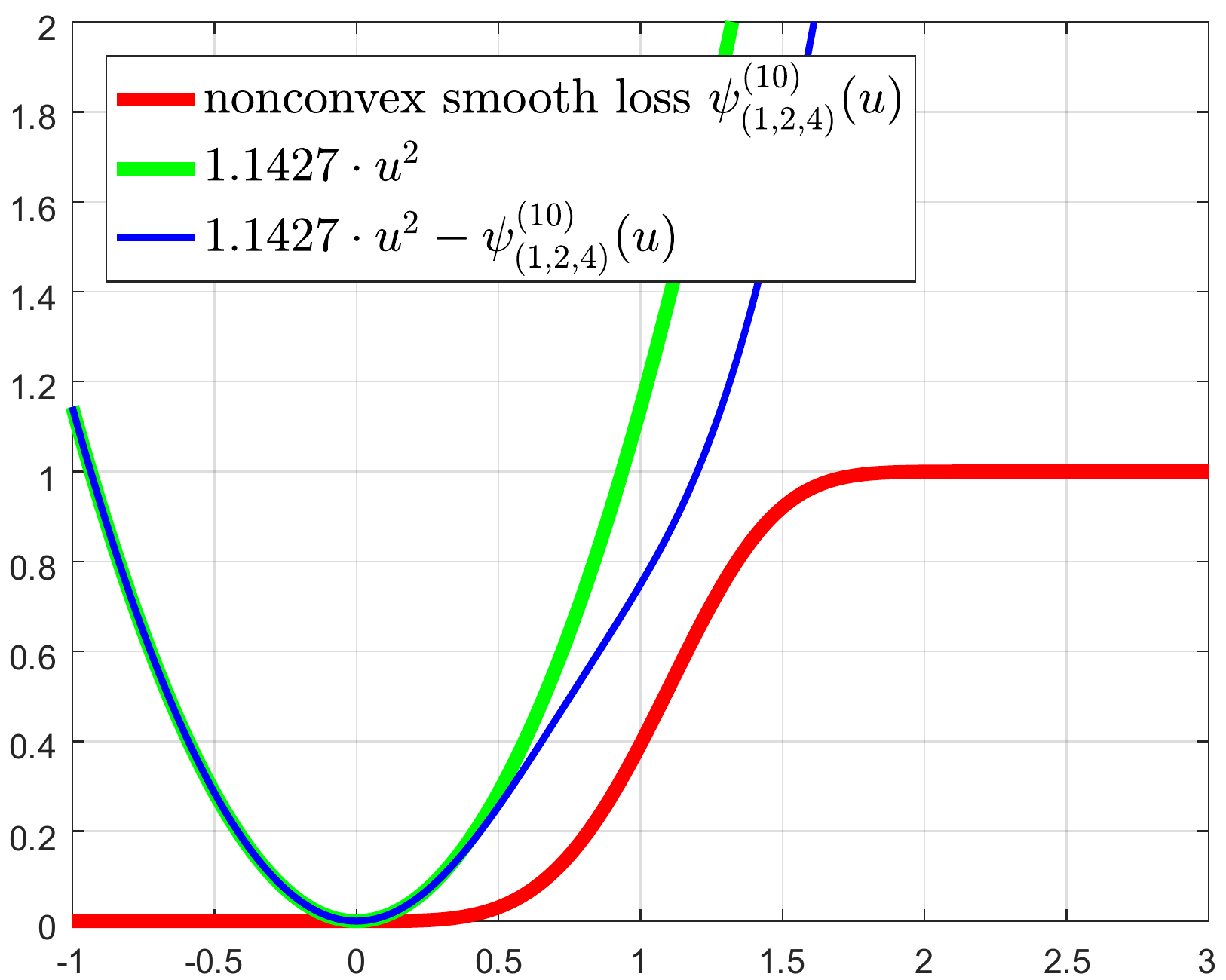}
}
\caption{The plots of some LS-DC losses for classification and their DC decompositions: ``\rm{\crd \textbf{Red~curve}}= {\cgr \textbf{Green~curve}} - {\cbl \textbf{Blue~curve}}".
In the plot, the {\textbf{Black curve}} (if it exists) is the plot of the original non-LS-DC loss which is approximated by an LS-DC loss({\crd red curve}), the {\cgr \textbf{Green~curve}} is the function $Au^2$ and the {\cbl \textbf{Blue~curve}} is the convex function $Au^2-\psi(u)$. The loss names in the legends are defined in Table \ref{tab:classification_loss} in Appendix \ref{appendix_a}. All of the LS-DC parameters $A$ are chosen as the lower bounds in Table \ref{tab:classification_loss}, and increasing the value of $A$ can make the {\cbl Blue curve} "\textbf{smoother}".
}\label{fig_lossC}
\end{figure}

\begin{proposition}[LS-DC property of regression losses]\label{subsec:regr_loss}
The commonly used regression losses are LS-DC loss or can be approximated by LS-DC loss. We enumerate them as follows.
\begin{enumerate}[(1)]
  \item The least squares loss and the truncated least squares loss are all LS-DC losses with $A\geq1$.
  \item The $\varepsilon$-insensitive loss $\ell_\varepsilon(y,t):= (|y-t|-\varepsilon)_+$, mostly used for SVR, is \textbf{not} an LS-DC loss. However, we can smooth it as
      \begin{equation}\label{eq:smooth_insensitive loss}
          \ell_{(\varepsilon,p)}(y,t):=\tfrac{1}{p}\log(1+\exp(-p(y-t+\varepsilon))) + \tfrac{1}{p}\log(1+\exp(p(y-t-\varepsilon))),
      \end{equation}
      which is LS-DC loss with $A\geq p/4$.
    \item The absolute loss $\ell(y,t) = |y-t|$ is also \textbf{not} an LS-DC loss. However, it can be smoothed by LS-DC losses. For instance, Hubber loss $$\ell_\delta(y,t) =
        \left\{\begin{array}{ll}
            \frac1{2\delta}(y-t)^2, & |y-t|\leq\delta, \\
             |y-t|-\frac{\delta}{2}, & |y-t|>\delta,
           \end{array}\right.
        $$
        which approximates the absolute loss is an LS-DC loss with $A\geq 1/(2\delta)$; Setting $\varepsilon=0$ in \eqref{eq:smooth_insensitive loss}, we obtain another smoothed absolute loss, which is an LS-DC loss with $A\geq p/4$.
    \item The truncated absolute loss $\ell_a(y,t):=\min\{|y-t|,a\}$ can be approximated by the truncated Hubber loss $\min\{\ell_\delta(y,t),a\}$, which is an LS-DC loss with $A\geq 1/(2\delta)$.
\end{enumerate}
\end{proposition}

Some regression losses and their DC decompositions are plotted in Figure \ref{fig_lossR}.
\begin{figure}[htbp]
\centering
\subfigure[Smoothed $\varepsilon$-insensitive loss]{
\includegraphics[width=0.31\textwidth]{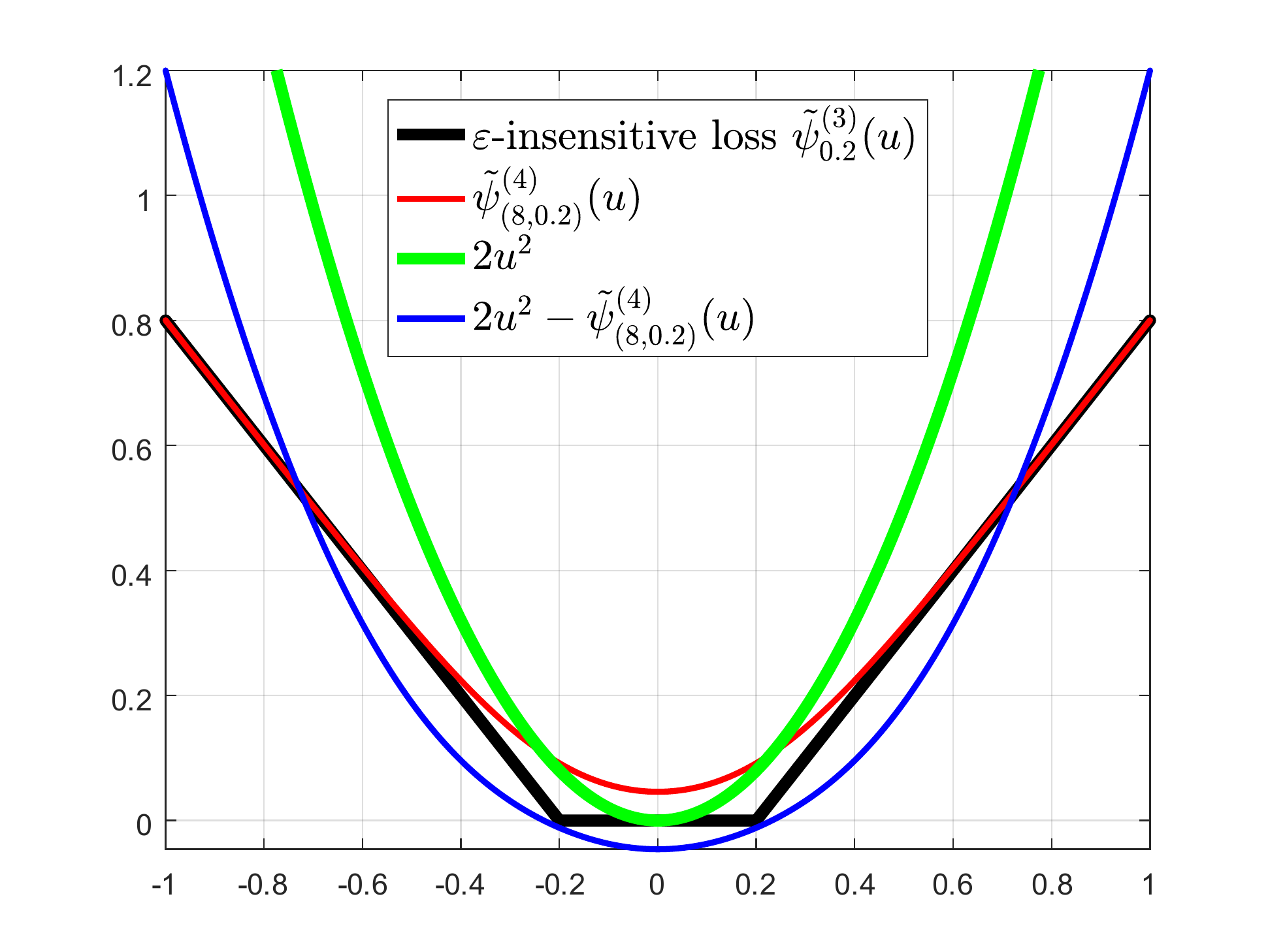}
}~
\subfigure[Absolute and Hubber loss]{
\includegraphics[width=0.31\textwidth]{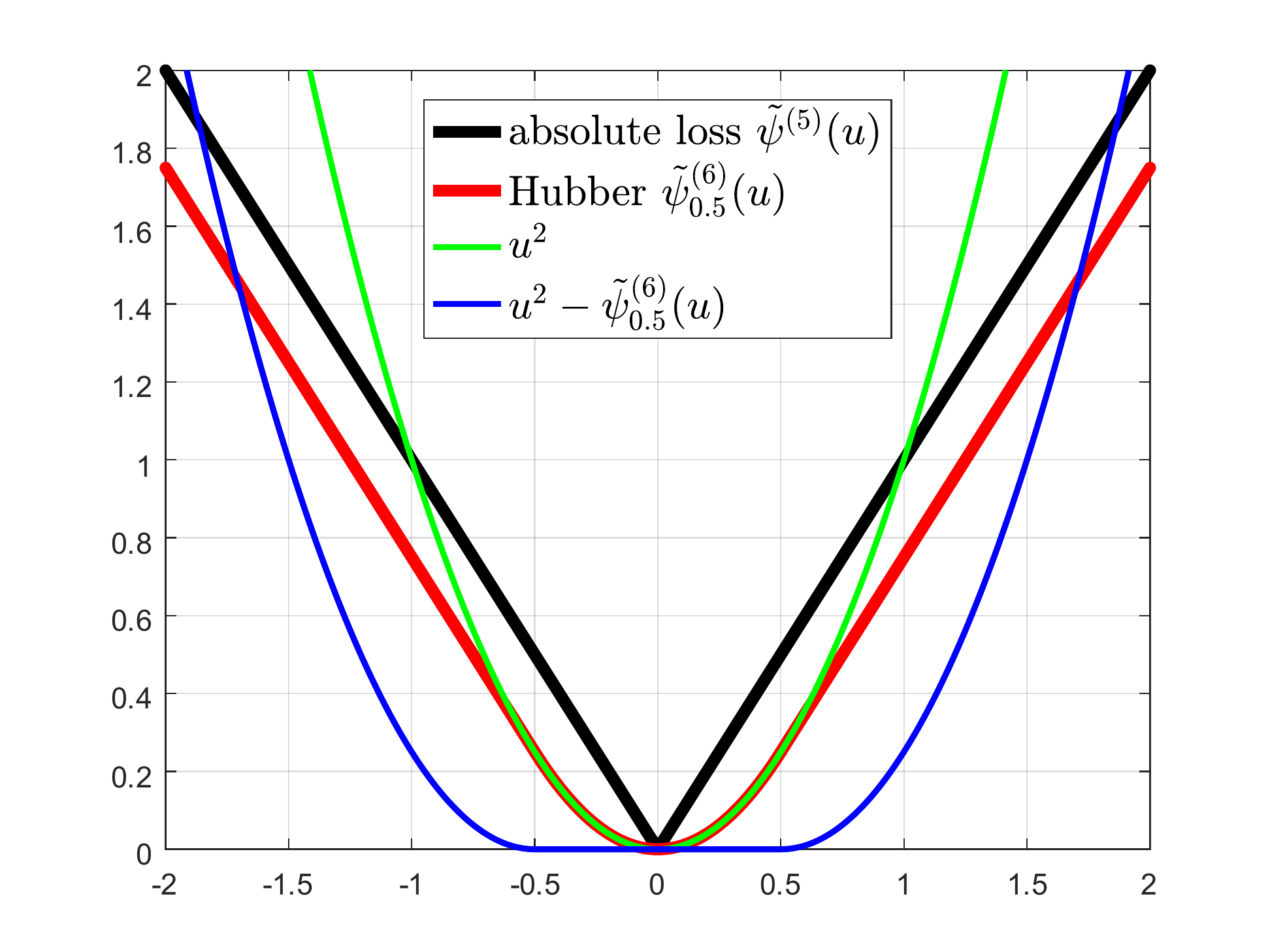}
}~
\subfigure[Truncated Hubber loss]{
\includegraphics[width=0.31\textwidth]{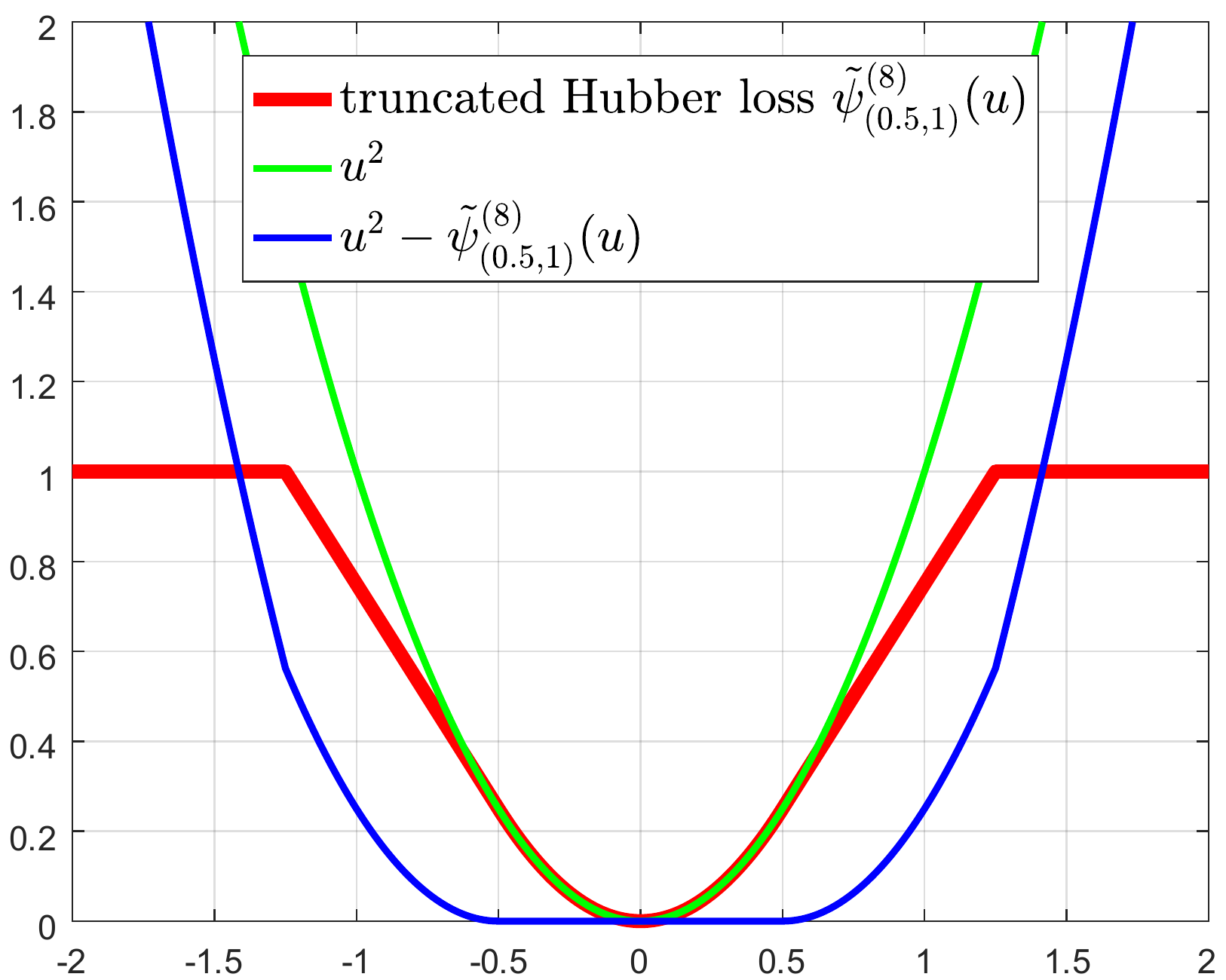}
}
\caption{The plots of some LS-DC losses for regression and their DC decompositions:
$``\rm{\crd \textbf{Red~curve}}= {\cgr \textbf{Green~curve}} - {\cbl \textbf{Blue~curve}}"$.
In the plot, the {\textbf{Black curve}} (if it exists) is the plot of the original non-LS-DC loss which is approximated by an LS-DC loss({\crd red curve}), the {\cgr \textbf{Green~curve}} is the function $Au^2$ and the {\cbl \textbf{Blue~curve}} is the convex function $Au^2-\psi(u)$. The loss names in the legends are defined in Table \ref{tab:classification_loss} in Appendix \ref{appendix_a}. All the LS-DC parameters $A$ are chosen as the lower bounds in Table \ref{tab:classification_loss}.
}\label{fig_lossR}
\end{figure}

\section{Unified algorithm for SVM models with LS-DC losses}\label{sec:UniSVM}

Let $\ell(y,t)$ be any LS-DC loss discussed in Section \ref{sec:lsdc_loss}, and let $\psi(u)$ satisfying $\psi(1-yt)=\ell(y,t)$ (for classification task) or $\psi(y-t)=\ell(y,t)$ (for regression task) have the DC decomposition \eqref{eq:lsdc_loss} with parameter $A>0$. The SVM model \eqref{eq:svm_a} with any loss can then be decomposed as
\begin{equation}\label{eq:svm_dc}
  \min_{\valpha\in\Real^m}\lambda \valpha^\top \K\valpha + \frac A m \|\y-\K\valpha\|^2-\left(\frac A m\|\y-\K\valpha\|^2 -\frac1m\sum_{i=1}^{m}\psi\left(r_i\right)\right),
\end{equation}
where $r_i = 1-y_i\K_i\valpha$ (for classification) and $r_i = y_i - \K_i\valpha$ (for regression).

Owing to the DCA procedure \eqref{eq:dca}, with an initial point $\valpha^0$, a stationary point of \eqref{eq:svm_dc} can be iteratively reached by solving

\begin{equation}\label{eq:dcrsvc}
  \valpha^{k+1}\in\argmin_{\valpha\in\Real^m} \lambda\valpha^\top\K\valpha + \tfrac A m \|\y-\K\valpha\|^2 + \left\langle \tfrac{2}{m}\K\left(A(\y-\vxi^k) {-} \vgamma^k\right), \valpha\right\rangle,
\end{equation}
where $\vxi^k = \K\valpha^k$ and  $\vgamma^k = (\gamma^k_1,\gamma^k_2,\cdots, \gamma^k_m)^\top$ satisfies
{\begin{equation}\label{eq:gamma_update}
  \gamma_i^k \in  \tfrac12 y_i \partial\psi(1-y_i\vxi^k_i) {\rm{(classication)~or~~}} \gamma_i^k \in  \tfrac12 \partial\psi(y_i-\vxi^k_i) {\rm{(regression),}}
\end{equation}
where $\partial\psi(u)$ indicates the subdifferential of the convex function $\psi(u)$.} The related losses and their derivatives or subdifferentials for updating $\vgamma^k$ in \eqref{eq:gamma_update} are listed in Table \ref{tab:classification_loss} in Appendix \ref{appendix_a}.

The KKT conditions of \eqref{eq:dcrsvc} are
\begin{equation}\label{eq:kkt_dcrsvc}
  \left(\tfrac{\lambda m}{A}\K +\K\K^\top\right)\valpha =  \K(\vxi^k- \tfrac{1}{A}\vgamma^k).
\end{equation}
By solving \eqref{eq:kkt_dcrsvc}, we propose a unified algorithm that can train SVM models with any LS-DC loss. For different LS-DC losses (either classification loss or regression loss), we just need to calculate the different $\vgamma$ by \eqref{eq:gamma_update}.
We address the algorithm as \textbf{UniSVM}, which is summarized as Algorithm \ref{alg1}.

\begin{algorithm}[htp]
\caption{\textbf{UniSVM}(\textbf{Uni}fied \textbf{SVM})}\label{alg1}
\renewcommand{\algorithmicrequire}{\textbf{Input:}}
\renewcommand\algorithmicensure {\textbf{Output:} }
\begin{algorithmic}[1] %
\REQUIRE Given a training set $\sT=\{(\x_i, y_i)\}_{i=1}^m$ with $\x_i\in\Real^d$ and $y_i\in\{-1, +1\}$  or $y_i\in\Real$; Kernel matrix $\K$ satisfying $\K_{i,j}=\kappa(\x_i,\x_j)$, or $\PM$ satisfying $\PM\PM^\top\approx \K$ satisfying $\PM_\sB\PM^\top=\K_\sB$; Any LS-DC loss function $\psi(u)$ with parameter $A>0$; The regularizer $\lambda$.
\ENSURE The prediction function $f(\x)=\sum_{i=1}^{m}\alpha_i \kappa(\x_i,\x)$ with $\valpha = \valpha^k$.\\
\STATE $\vgamma^0=0$, $\vxi^0=\y$; Set $k:=0$.
\WHILE {not convergence}
\STATE Solving \eqref{eq:kkt_dcrsvc} with respect to \eqref{eq:alpha_small}, \eqref{eq:full_low_update} or \eqref{eq:alpha_sparse} to obtain $\valpha^{k+1}$, where the inversion is only calculated in the first iteration;
\STATE Update $\vxi^{k+1}=\K\valpha^{k+1}$ or $\vxi^{k+1} = \PM\PM_\sB^\top \valpha_\sB^{k+1}$, $\vgamma^{k+1}$ by \eqref{eq:gamma_update}; $k:=k+1.$
\ENDWHILE
\end{algorithmic}
\end{algorithm}

The new algorithm possesses the following advantages:
\begin{itemize}
  \item[$\bullet$] It is suitable for training any kind of SVM models with any LS-DC losses, including convex loss or nonconvex loss. The training process for classification problems is also the same as for regression problems. The proposed UniSVM is therefore definitely a unified algorithm.
  \item[$\bullet$] For nonconvex loss, unlike the existing algorithms \citep{Collobert2006,Wu_Liu2007,Tao2018,Feng2016} that must iteratively solve L1SVM/L2SVM or reweighted L2SVM in the inner loops, UniSVM is free of the inner loop because it solves a system of linear equations \eqref{eq:kkt_dcrsvc} with a closed-form solution per iteration.

  \item[$\bullet$] According to the studies on LSSVM in \cite{Zhou2016}, the problem \eqref{eq:kkt_dcrsvc} may have multiple solutions, including some sparse solutions, if $\K$ has low rank\footnote{$\K$ is always low rank in computing, for there are always many similar samples in the training set, leading the corresponding columns of the kernel matrix to be (nearly) linearly dependent.}. This is of vital importance for training large-scale problems efficiently. Details will be discussed in subsection \ref{sec:sub_largecase}.

  \item[$\bullet$] In experiments, we always set $\vxi^0 =\y$ and $\vgamma^0=0$ instead of giving an $\valpha^0$ to begin the algorithm. This is equivalent to starting the algorithm from the solution of LSSVM, which is a moderate guess of the initial point, even for nonconvex loss.
\end{itemize}

In subsection \ref{sec:sub_smallcase}, we present an easily grasped version for the proposed \textbf{UniSVM} in the case that the full kernel $\K$ is available. In subsection \ref{sec:sub_largecase}, we propose an efficient method to solve the KKT conditions \eqref{eq:kkt_dcrsvc} for \textbf{UniSVM} even if the full kernel matrix is unavailable. The Matlab code is also given in Appendix \ref{appendix:smallcase}.

\subsection{Solving UniSVM with full kernel matrix available}\label{sec:sub_smallcase}

If the full kernel matrix $\K$ is available and $\tfrac{\lambda m}{A}\I + \K$ can be inverted cheaply, then noting $\Q=\left(\tfrac{\lambda m}{A}\I +\K\right)^{-1}$, we can prove that
\begin{equation}\label{eq:alpha_small}
 \valpha^{k+1} = \Q(\vxi^k -\tfrac1A \vgamma^k)
 \end{equation}
is one nonsparse solution of \eqref{eq:kkt_dcrsvc}. It should be noted that $\Q$ is only calculated once. Hence, after the first iteration, $\valpha^{k+1}$ will be reached within $O(m^2)$.

Furthermore, if $\K$ is low rank and can be factorized as $\K=\PM\PM^\top$ with a full-column rank $\PM\in\Real^{m\times r}$ ($r<m$), the cost of the process can be reduced through two skillful methods. One is SMW identity \citep{Golub1996}, which determines the cost $O(mr^2)$ to compute $\PM^\top \PM$, the cost $O(r^3)$ to obtain the inversion $\hat\Q=\left(\tfrac{\lambda m}{A}\I +\PM^\top \PM\right)^{-1}\in\Real^{r\times r}$  once, and the cost within $O(mr)$ to update the nonsparse $\valpha^{k+1}$ per iteration as
\begin{equation}\label{eq:full_low_update}
  \valpha^{k+1} = \tfrac{A}{\lambda m}\left(\I - \PM {\hat\Q} \PM^\top\right)(\vxi^k - \tfrac1A\vgamma^k).
\end{equation}
The other is the method employed in subsection \ref{sec:sub_largecase} to obtain a sparse solution of \eqref{eq:kkt_dcrsvc}.

\subsection{Solving UniSVM for large-scale training with a sparse solution} \label{sec:sub_largecase}
For large-scale problems, the full kernel matrix $\K$ is always unavailable because of the limited memory and the computational complexity. Hence, we should manage to obtain the sparse solution of the model, since in this case $\K$ is always low rank or can be approximated by a low-rank matrix.

To obtain the low-rank approximation of $\K$, we can use the Nystr\"{o}m approximation \citep{Sun2015}, a kind of random sampling method, or the pivoted Cholesky factorization method proposed in \cite{Zhou2016} that has a guarantee to minimize the trace norm of the approximation error greedily.
The gaining approximation of $\K$ is $\PM\PM^\top$, where $\PM=[\PM_\sB^\top~~\PM_\sN^\top]^\top$
is a full column rank matrix {with $\PM_\sB \in \Real^{r\times r}$ ($r\ll m$)} and $\sB\subset\{1,2, \cdots,m\}$  being the index set corresponding to the only visited $r$ columns of $\K$.
Both algorithms satisfy the condition that the total computational complexity is within $O(mr^2)$, and $\K_\sB$, the visited rows of $\K$ corresponding to set $\sB$, can be reproduced exactly as $\PM_\sB\PM^\top$.

Replacing $\K$ with $\PM\PM^\top$ in \eqref{eq:kkt_dcrsvc}, we have $$
  \PM(\tfrac{\lambda m}{A}\I + \PM^\top\PM)\PM^\top\valpha =  \PM(\PM^\top(\vxi^k - \tfrac1A \vgamma^k)),$$
which can be simplified as
\begin{equation}\label{eq:app2}
 \left(\tfrac{\lambda m}{A}\I + \PM^\top\PM\right)\PM^\top\valpha = \PM^\top(\vxi^k -\tfrac1A \vgamma^k).
\end{equation}
This is because $\PM$ is a full column rank matrix. 
By simple linear algebra, if we let $\valpha=[\valpha_\sB^\top ~~\valpha_\sN^\top]^\top$ be a partition of $\valpha$ corresponding to the partition of $\PM$, then we can set $\valpha_\sN=0$ to solve \eqref{eq:app2}. Thus, \eqref{eq:app2} is equivalent to $$
 \left(\tfrac{\lambda m}{A}\I + \PM^\top\PM\right)\PM_\sB^\top\valpha_\sB = \PM^\top(\vxi^k -\tfrac1A \vgamma^k).
$$
We then have
\begin{equation}\label{eq:alpha_sparse}
\valpha_\sB^{k+1} =\overline\Q \PM^\top(\vxi^k - \tfrac1A \vgamma^k).
\end{equation}
where $\overline\Q=\left((\tfrac{\lambda m}{A}\I + \PM^\top\PM)\PM_\sB^\top\right)^{-1}$, $\vxi^k = \PM\PM_\sB^\top \valpha_\sB^k$, and $\vgamma^k$ is updated by \eqref{eq:gamma_update}.

Notice that $\overline\Q$ is only calculated in the first iteration with the cost $O(r^3)$. The cost of the algorithm is $O(mr^2)$ for the first iteration, and $O(mr)$ for the following iterations. Hence, UniSVM can be run very efficiently.

\section{Experimental Studies}\label{sec:exp}

In this section, we present some experimental results to illustrate the effectiveness of the proposed unified model. All the experiments are run on a computer with an Intel Core i5-6500 CPU @3.20GHz$\times4$ and a maximum memory of 8GB for all processes; the computer runs Windows 7 with Matlab R2016b. The comparators include L1SVM and SVR solved by \texttt{LibSVM}, L2SVM and the robust SVM modes in \citet{Collobert2006, Tao2018, Feng2016}.

\subsection{Intuitive comparison of UniSVM with other SVM models on small data sets}
In this subsection, we first present experiments to show that the proposed UniSVM with convex loss can obtain the comparable performance by solving L1SVM, L2SVM and SVR on the small data sets.
Second, we perform experiments to illustrate that UniSVM with nonconvex loss also more efficiently obtains comparable performance by solving some robust SVMs with nonconvex loss compared with the algorithms in \citet{Collobert2006, Tao2018, Feng2016}. We have implemented UniSVM in two cases; one is in \eqref{eq:alpha_small} with the full kernel matrix $\K$ available, called \textbf{UniSVM-full}, and the other is to obtain the sparse solution of the model by \eqref{eq:alpha_sparse}, where $\K$ is approximated as $\PM\PM^\top$ with $\PM\in\Real^{m\times r} (r\ll m)$, noted as \textbf{UniSVM-app}. The latter has the potential to resolve the large-scale tasks.
L1SVM and SVR are solved by the efficient tools \texttt{LibSVM} \citep{Chang2011}, and the other related models (L2SVM, the robust L1SVM and the robust L2SVM) are solved by the solver of quadratic programming \texttt{quadprog.m} in Matlab.

\subsubsection{On convex loss cases}

The first experiment is a hard classification task on the highly nonlinearly separable ``xor" data set as Fig. \ref{fig_classplot} shows, where the instances are generated by uniform sampling with 400 training samples and 400 test samples. The kernel function is $\kappa(\x,\z) = \exp(-\gamma\|\x-\z\|^2)$ with $\gamma = 2^{-1}$, $\lambda$ is set as $10^{-5}$ and $r=10$ for UniSVM-app. The experimental results are plotted in Fig. \ref{fig_classplot} and the detailed information of the experiments is given as the captions and the subtitles of the figures.

\begin{figure}[htp]
  \centering
  \includegraphics[width=\textwidth]{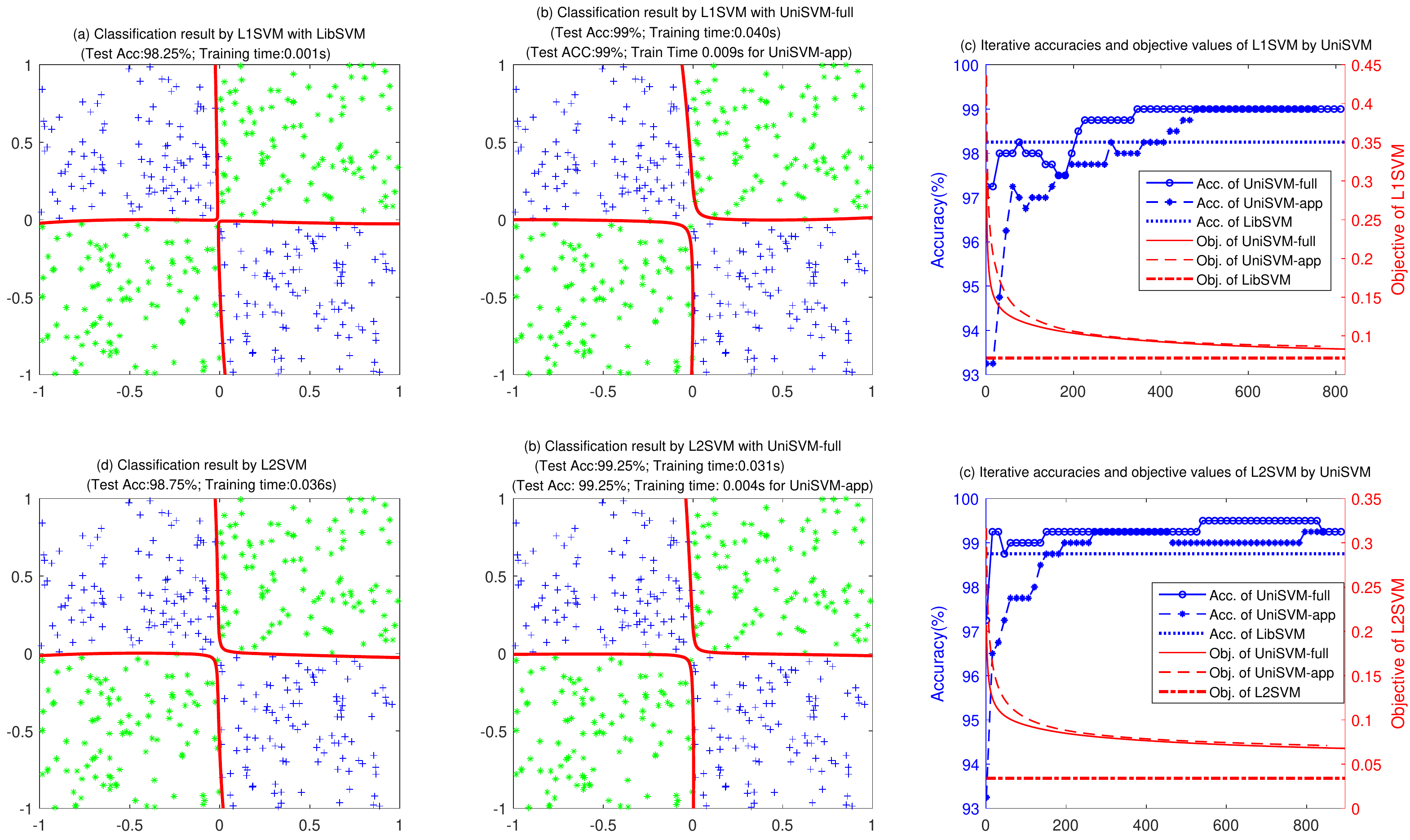}
  \caption{
  Comparison of the related algorithms of SVM with convex losses. In (a), (b), (d) and (e), the classification results of the algorithms are plotted as red solid curves (since the differences between them are slight, the classification curves of UniSVM-app are not plotted and its test accuracies and the training times are correspondingly noted in (b) and (e)).
   In (c) and (f), the iterative processes of UniSVM are plotted. The blue curves with respect to the left y-axis are the iterative test accuracies of UniSVM, and the red curves with respect to the right y-axis are the iterative objective values of \eqref{eq:svm_a}. The accuracies and the objective values of L1SVM/L2SVM are plotted as the horizontal lines for reference.
  }\label{fig_classplot}
\end{figure}

From the experimental results in Fig. \ref{fig_classplot}, we have the following findings:
\begin{itemize}
  \item The proposed UniSVM for solving L1SVM or L2SVM can obtain similar performance compared with the state-of-the-art algorithms (LibSVM/quadporg). Of course, in those smaller cases, LibSVM is more efficient than UniSVM, since they are only designed for SVM with convex losses.
  \item The low-rank approximation of the kernel matrix can significantly accelerate UniSVM, and the acceleration rate is approximated as $\frac mr$.
  \item The red curves in Fig. \ref{fig_classplot}(c) and (f) reveal that UniSVM is the majorization-minimization algorithm \citep{Naderi2019}. All cases of UniSVM reach the optimal value of L1SVM or L2SVM from above. In this setting, if $r>15$, the difference of the objective values between UniSVM-full and UniSVM-app will vanish.
\end{itemize}
Thus, the advantage of UniSVM lies in solving the large-scale tasks with low-rank approximation.

The second set of experiments is based on a regression problem based on an SVR model \eqref{eq:svr} with $\varepsilon$-insensitive loss, where 1,500 training samples and 1,014 test samples are generated by the Sinc function with Gaussian noise $y = \frac{\sin(x)}{x} + \zeta$ with $x\in[-4\pi, 4\pi]$ discretized by the step of 0.01 and $\zeta\sim N(0,0.05)$. Since the $\varepsilon$-insensitive loss is not LS-DC loss, we compare LibSVM with UniSVM with smoothed $\varepsilon$-insensitive loss \eqref{eq:smooth_insensitive loss} for solving SVR \eqref{eq:svr}. Here, the kernel function $\kappa(\x,\z) = \exp(-\gamma\|\x-\z\|^2)$ with $\gamma = 0.5$, $\lambda$ is set as $10^{-4}$ and $r=50$ for UniSVM-app. The experimental results are plotted in Fig. \ref{fig_regplot}.
\begin{figure}[htp]
  \centering
  \includegraphics[width=\textwidth]{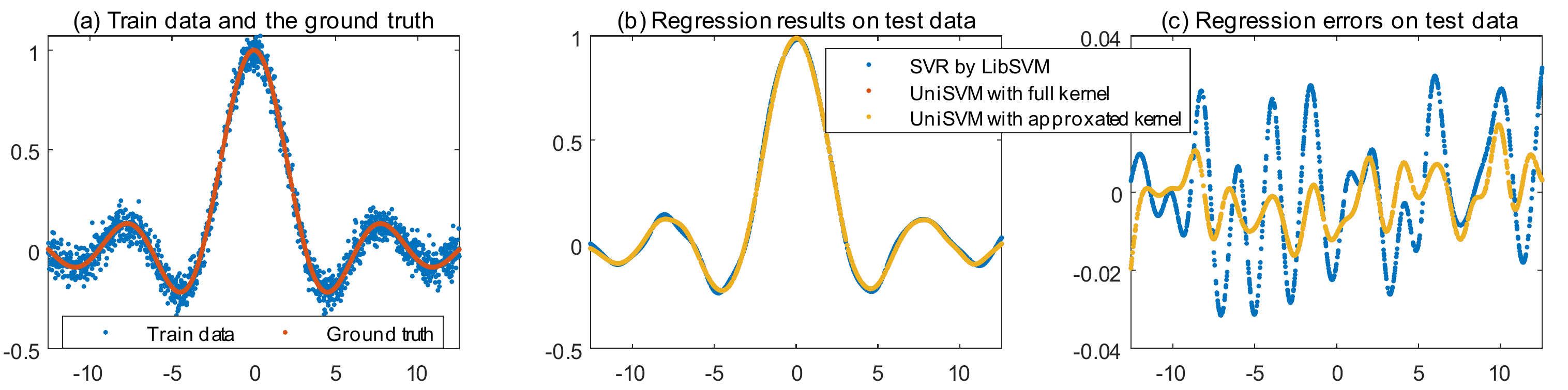}
  \caption{The comparison of solving SVR by UniSVM with smooth $\varepsilon$-insensitive loss and LibSVM on Sinc regression problem. In (a), the training data and the ground truth are plotted; in (b), the regression results of LibSVM, UniSVM-full and UniSVM-app are given, where the respective MSRs are 0.0027, 0.0025, and 0.0025, and the training times are 0.062s, 0.144s and 0.035s; in (c), the regression errors of three algorithms are plotted. In (b) and (c), the difference between UniSVM-full and UniSVM-app can be neglected, while the training time of the latter is less than one-fifth of that of the former.
  }\label{fig_regplot}
\end{figure}

From the experimental results in Fig. \ref{fig_regplot}, we have two findings. One is that the new UniSVM can simply achieve better performance than LibSVM. This may be because the added noise follows a Gaussian distribution while UniSVM is initialized as LSSVM. The other finding is that the low-rank approximation of the kernel matrix is highly efficient here, since UniSVM-app can obtain results similar to those of UniSVM-full, which reveals the similar findings as in \cite{Zhou2016, Chen2018}. The speedup rate here is less than $\frac mr$, which is because the number of iterations of UniSVM is only 3.

\subsubsection{On nonconvex loss cases}
The first set of experiments was again performed on the ``xor'' data set as shown in Fig. \ref{fig_regplot}, where some training samples (10\%) are contaminated to simulate the outliers by flipping of their labels (the test samples are noise-free). The compared algorithms include robust L1SVM in \cite{Collobert2006}, MS-SVM of robust L2SVM in \cite{Tao2018}, and the re-weighted L2SVM in \cite{Feng2016}. The results are presented in Fig. \ref{fig_class_robust_plot}, in which the classification results of L1SVM and L2SVM are listed as the references. In these experiments, only UniSVM was implemented as UniSVM-app with $r=10$, shortened as UniSVM. The truncated parameter $a=2$ for nonconvex loss.
\begin{figure}[htp]
  \centering
  \includegraphics[width=\textwidth]{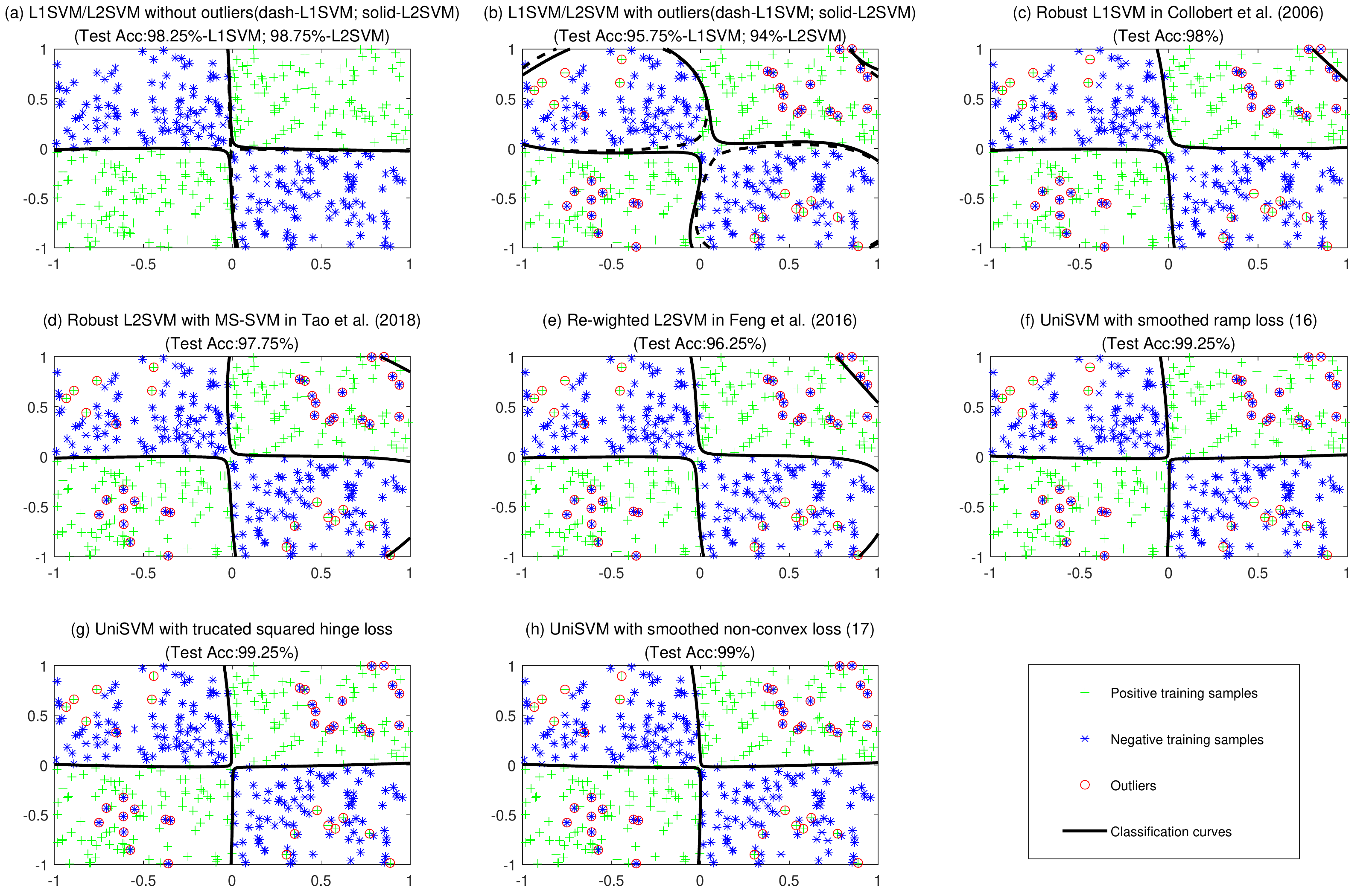}
  \caption{Comparison of the classification results of the related algorithms of SVM with convex and nonconvex losses. See the titles of the subfigures for details, where the used loss \eqref{eq:los4} in figure (h) with $a=b=c=2$ is just the same as \eqref{eq:los2} used in figure (e). It is worth noting that the results of (d) and (g) are based on the same model with the truncated squared hinge loss with the different algorithms, and (f) and (i) are based on the same model with nonconvex smooth loss \eqref{eq:los2} but are solved with a different algorithm.
  }\label{fig_class_robust_plot}
\end{figure}

From the experimental results in Fig. \ref{fig_class_robust_plot}, we observe that the models based on nonconvex loss do upgrade the classification results in cases of outliers. The algorithms in Fig. \ref{fig_class_robust_plot} (c)-(e), which need to iteratively solve L1SVM or L2SVM many times, are also affected by the outliers of the upper-right corner, where the outliers are dominated locally. However, the proposed UniSVM based on the effective LS-DC loss can completely resolve this problem. In particular, as highlighted by the results in Fig. \ref{fig_class_robust_plot}(d) and (g) where the same SVM model with truncated squared hinge loss is solved by different algorithms, the proposed UniSVM can solve the robust SVM with high performance. The comparison between the results of Fig. \ref{fig_class_robust_plot}(e) and (h) reveals a similar fact. The reason for this may be because the proposed UniSVM can obtain a better local minimum with a good initial point (a sparse solution of LSSVM) based on the DC decomposition of the corresponding nonconvex loss.

The second set of experiments is performed to compare the effectiveness of the related algorithms, also on the ``xor" problem. The training samples are randomly generated with varied sizes from 400 to 10,000, and the test samples are generated similarly with the same sizes. The training data is contaminated by randomly flipping the labels of 10\% of instances to simulate the outliers. We set $r=0.1m$ for all UniSVM algorithms to approximate the kernel matrix, and the other parameters are set as in the former experiments. The corresponding training time and test accuracies (averaged over ten trials) of the related robust SVM algorithms are plotted in Fig. \ref{fig_class_robust2}, where the results of LibSVM (with outliers and without outliers) are also given as references.
\begin{figure}
  \centering
  \includegraphics[width=\textwidth]{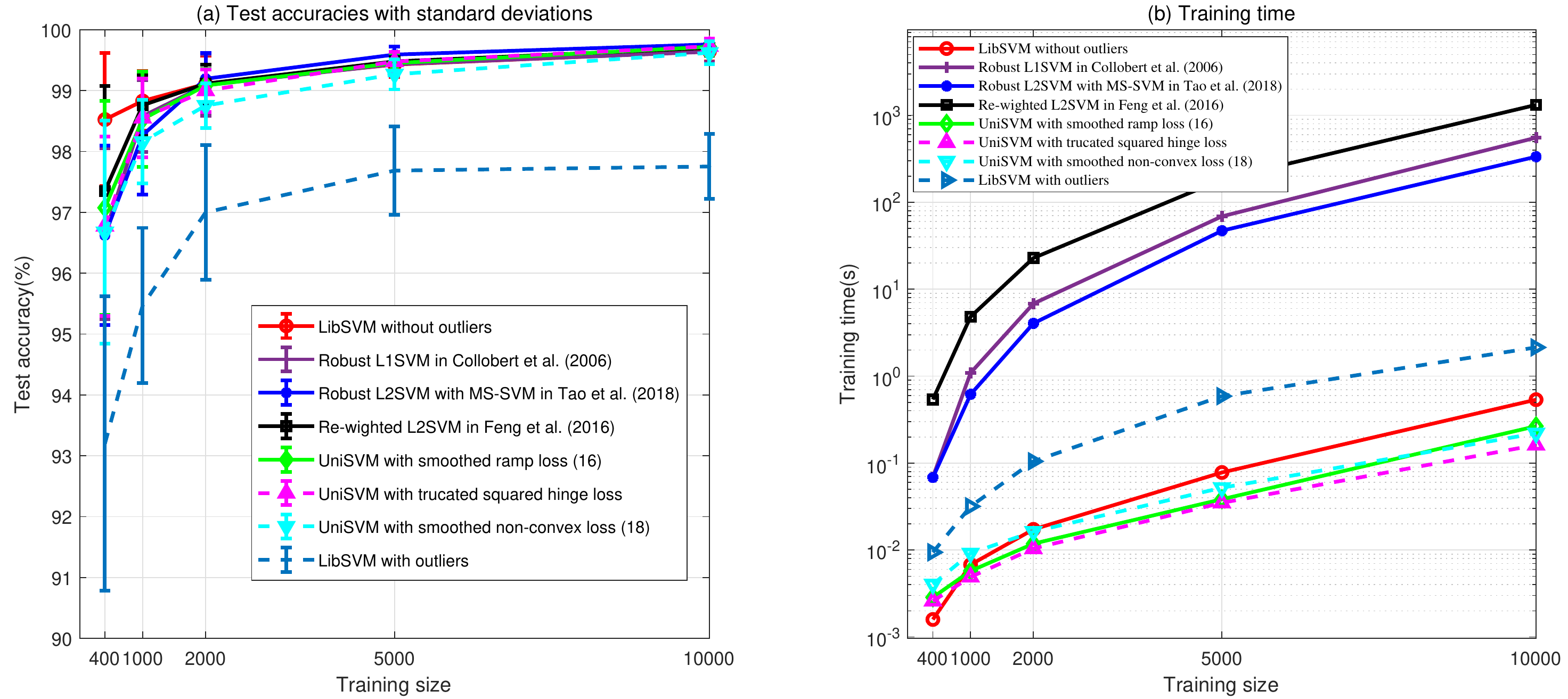}
  \caption{The comparisons of the training time and test accuracies (averaged over ten trials) of the related algorithms on the contaminated ``xor" data sets with different training size. }\label{fig_class_robust2}
\end{figure}

From the experimental results in Fig. \ref{fig_class_robust2}, we have the following findings:
\begin{itemize}
  \item In Fig. \ref{fig_class_robust2}(a), it is clear that the performances of all the related robust SVM algorithms based on nonconvex losses are better than that of the L1SVM with convex loss on the contaminated training data sets and that all of them match the results of the noise-free case. At the same time, we notice that the differences of the test accuracies between the selected robust algorithms are very small.
  \item From Fig. \ref{fig_class_robust2}(b), we observe that the differences of the training time between the related algorithms are large, especially for the larger training set. The training time of the proposed UniSVM is significantly less, while the robust L1SVM \citep{Collobert2006}, Robust L2SVM \citep{Tao2018} and re-weighted L2SVM \citep{Feng2016}, which need to solve the constrained QP several times, have long training times. Even they can be more efficiently implemented (such as SMO) than \texttt{quadprog.m}, but their training time will be longer than that of LibSVM since all of them at least solve a QP which is similar to the QP solved by LibSVM.
  \item In our setting, the outliers greatly affect the results of LibSVM not only for test accuracies but also training time.
   \item All in all, the new proposed UniSVM with low rank kernel approximation is not only robust to noise but also highly efficient with respect to the training process.
\end{itemize}

\subsection{Experiments on larger benchmark data sets}
In this section, we perform experiments to show that UniSVM can quickly train the convex and nonconvex SVM models with comparable performance using a unified scheme on large data sets. We choose only the state-of-the-art SVM tool \textbf{LibSVM} \citep{Chang2011}(including SVC and SVR) as the comparator, rather than other robust SVM algorithms in previous papers \citep{Collobert2006, Wu_Liu2007,Tao2018, Feng2016}, to conserve experimental time.

First, we select 4 classification tasks and 3 regression tasks from the UCI database to illustrate the performance of the related algorithms.
The detailed information of the data sets and the hyper-parameters (\texttt{training size, test size, dimension, $\lambda$, $\gamma$}) is given as follows:\\
\resizebox{0.98\textwidth}{!}{
\begin{tabular}{rlrl}
  \textbf{Adult:}&(\texttt{32561, 16281, 123, $10^{-5}$, $2^{-10}$}), &
  \textbf{Ijcnn:}&(\texttt{49990, 91790, 22,$10^{-5}$, $2^{0}$}), \\
  \textbf{Shuttle:}&(\texttt{43500, 14500, 9, $10^{-5}$, $2^{1}$}), &
  \textbf{Vechile:}&(\texttt{78823, 19705, 100, $10^{-2}$, $2^{-3}$}); \\
  \textbf{Cadata:}&(\texttt{10640, 10000, 8, $10^{-2}$, $2^{0}$}),  &
  \textbf{3D-Spatial:}&(\texttt{234874, 200000, 3, $10^{-3}$, $2^{6}$}), \\
  \textbf{Slice:}&(\texttt{43500, 10000, 385, $10^{-9}$, $2^{-5}$}). & &  \\
\end{tabular}}\\

Here, the classification tasks have the default splitting, and the regression tasks are split randomly.
The $\lambda$ (regularizer) and $\gamma$ (for Gaussian kernel $\kappa(\x,\z) = \exp(-\gamma\|\x-\z\|^2)$) are roughly chosen by the grid search. For the parameters of loss functions, we simply use the default value (given next). The fine-tuning of all parameters will certainly improve the performance further.

To implement the UniSVM for larger training data, we use the pivoted Cholesky factorization method proposed in \citet{Zhou2016} to approximate the kernel matrix $\K$, and the low-rank approximation error is controlled by the first matched criterion $trace(\K-\tilde{\K})<0.001 \cdot m$ or $r\le1000$, where $m$ is the training size and $r$ is the upper bound of the rank.

The first set of experiments shows that the proposed UniSVM can train SVM models with convex or nonconvex loss for classification problems. The chosen losses for UniSVM$_1$ to UniSVM$_{10}$ are listed as following:

\resizebox{0.95\textwidth}{!}{
\begin{tabular}{ll}
  UniSVM$_1$:  least squares loss, &~UniSVM$_2$:  smoothed hinge ($p=10$),\\
  UniSVM$_3$:  squared hinge loss, &~UniSVM$_4$:  truncated squared hinge ($a=2$),\\
  UniSVM$_5$:  truncated least squares ($a=2$), &~UniSVM$_6$:  loss \eqref{eq:ramp_s} ($a=2$),\\
  UniSVM$_7$:  loss \eqref{eq:ramp_sp} ($p=10$), &~UniSVM$_8$:  loss \eqref{eq:los4} ($a=b=c=2$),\\
  UniSVM$_9$:  loss \eqref{eq:los4} ($a=b=2, c=4$),  &UniSVM$_{10}$:  loss \eqref{eq:los4} ($a=2, b=3, c=4$).\\
\end{tabular}}
\vspace{0.2cm}

The results in Table \ref{exp:class} are obtained based on the original data sets, and those of Table \ref{exp:class2} are based on the contaminated data sets, where 20\% of the labels of training instances are randomly flipped. Since the kernel approximation in \cite{Zhou2016} undergoes random initialization, the training times recorded in Matlab are not very stable, and random flipping is observed for noise cases, all results are averaged over ten random trials.

\begin{table}[htp]
  \centering
  \caption{\textbf{Classification tasks I}--Test accuracies and the training times of the related algorithms on the benchmark data sets. All results are averaged over ten trials with the standard deviations in brackets; The first four lines are based on convex losses and the others are based on nonconvex losses.}\label{exp:class}
\resizebox{\textwidth}{!}{
\setlength{\tabcolsep}{1mm}
{
\begin{tabular}{rcccccccc}
\toprule
&\multicolumn{4}{c}{Test accuracy (\%)} & \multicolumn{4}{c}{Training time (CPU seconds)}\\
\cmidrule(r){2-5} \cmidrule(r){6-9}
Algorithm& Adult&Ijcnn&Shuttle&Vechile&Adult&Ijcnn&Shuttle&Vechile\\
\cmidrule(r){1-9}
{LibSVM~~~}&{84.65}({0.00)}&{98.40}({0.00)}&{99.81}({0.00)}&{84.40}({0.00)}&{51.49}({0.07)}&{23.02}({0.63)}&{4.75}({0.16)}&{1091}({175)}\\
{UniSVM$_1$}&{84.56}({0.02)}&{94.65}({0.07)}&{98.80}({0.04)}&{85.24}({0.01)}&{\textbf{0.44}}({0.02)}&{\textbf{18.24}}({0.15)}&{\textbf{0.34}}({0.02)}&{\textbf{34.77}}({0.44)}\\
{UniSVM$_2$}&{84.68}({0.04)}&{97.07}({0.05)}&{99.81}({0.01)}&{84.42}({0.00)}&{0.98}({0.04)}&{38.51}({0.32)}&{2.44}({0.10)}&{36.66}({0.62)}\\
{UniSVM$_3$}&{85.13}({0.02)}&{98.22}({0.03)}&{99.82}({0.00)}&{85.23}({0.01)}&{0.62}({0.03)}&{35.40}({0.28)}&{2.03}({0.06)}&{35.34}({0.39)}\\
\cmidrule(r){1-9}
{UniSVM$_4$}&{84.75}({0.04)}&{98.25}({0.04)}&{99.82}({0.00)}&{84.72}({0.00)}&{1.13}({0.04)}&{38.56}({0.53)}&{2.47}({0.06)}&{37.20}({0.42)}\\
{UniSVM$_5$}&{83.32}({0.05)}&{94.59}({0.08)}&{98.81}({0.04)}&{84.71}({0.00)}&{0.58}({0.03)}&{19.18}({0.35)}&{0.38}({0.02)}&{36.81}({0.52)}\\
{UniSVM$_6$}&{84.82}({0.02)}&{98.20}({0.03)}&{99.82}({0.00)}&{84.70}({0.00)}&{1.08}({0.03)}&{40.11}({0.44)}&{2.72}({0.10)}&{36.65}({0.34)}\\
{UniSVM$_7$}&{84.20}({0.02)}&{97.13}({0.05)}&{99.82}({0.00)}&{84.43}({0.00)}&{1.38}({0.04)}&{40.40}({0.51)}&{2.71}({0.06)}&{37.61}({0.35)}\\
{UniSVM$_8$}&{84.75}({0.02)}&{97.84}({0.04)}&{99.82}({0.00)}&{84.53}({0.00)}&{0.97}({0.04)}&{38.13}({0.26)}&{2.40}({0.11)}&{36.10}({0.45)}\\
{UniSVM$_9$}&{85.09}({0.02)}&{\textbf{98.46}}({0.04)}&{\textbf{99.83}}({0.00)}&{85.34}({0.00)}&{1.15}({0.05)}&{36.95}({0.29)}&{3.30}({0.11)}&{36.69}({0.44)}\\
{UniSVM$_{10}$}&{\textbf{85.16}}({0.03)}&{98.36}({0.03)}&{99.82}({0.00)}&{\textbf{85.49}}({0.00)}&{0.84}({0.03)}&{33.92}({0.22)}&{2.90}({0.15)}&{36.47}({0.55)}\\
\bottomrule
\end{tabular}}}
\end{table}

\begin{table}[htp]
  \centering
  \caption{\textbf{Classification tasks II}--Test accuracies and the training times of the related algorithms on the benchmark data sets with \emph{flipping of 20\% of training data labels}. All results are averaged over ten trials with the standard deviations in brackets; The first four lines are based on convex losses and the others are based on nonconvex losses.} \label{exp:class2}
\resizebox{\textwidth}{!}{
\setlength{\tabcolsep}{1mm}
{
\begin{tabular}{rcccccccc}
\toprule
&\multicolumn{4}{c}{Test accuracy (\%)} & \multicolumn{4}{c}{Training time (CPU seconds)}\\
\cmidrule(r){2-5} \cmidrule(r){6-9}
Algorithm& Adult&Ijcnn&Shuttle&Vechile&Adult&Ijcnn&Shuttle&Vechile\\
\cmidrule(r){1-9}
{LibSVM~~~}&{78.25}({0.00)}&{93.80}({0.00)}&{98.89}({0.00)}&{{84.28}({0.04)}
 }&{104.0}({0.8)}&{191.7}({1.2)}&{82.28}({1.52)}&{{1772}({123)}}\\
{UniSVM1}&{84.55}({0.02)}&{93.90}({0.08)}&{98.71}({0.04)}&{85.19}({0.00)}&{\textbf{0.44}}({0.06)}&{\textbf{18.34}}({0.33)}&{\textbf{0.35}}({0.02)}&{\textbf{34.76}}({0.43)}\\
{UniSVM2}&{82.27}({0.06)}&{93.72}({0.04)}&{99.01}({0.10)}&{84.25}({0.00)}&{0.65}({0.07)}&{20.56}({0.39)}&{0.60}({0.04)}&{36.99}({0.43)}\\
{UniSVM3}&{84.55}({0.02)}&{93.95}({0.08)}&{98.72}({0.04)}&{85.19}({0.00)}&{0.45}({0.06)}&{18.74}({0.34)}&{0.38}({0.02)}&{34.94}({0.44)}\\
\cmidrule(r){1-9}
{UniSVM4}&{84.26}({0.03)}&{97.36}({0.05)}&{99.81}({0.00)}&{84.37}({0.00)}&{1.23}({0.09)}&{40.23}({2.85)}&{2.52}({0.09)}&{37.53}({0.46)}\\
{UniSVM5}&{82.62}({0.03)}&{93.96}({0.03)}&{98.81}({0.02)}&{84.34}({0.00)}&{0.60}({0.05)}&{20.25}({0.48)}&{0.39}({0.02)}&{37.50}({0.48)}\\
{UniSVM6}&{84.25}({0.02)}&{97.59}({0.04)}&{99.81}({0.00)}&{84.46}({0.00)}&{1.14}({0.08)}&{48.74}({2.55)}&{2.72}({0.14)}&{37.19}({0.42)}\\
{UniSVM7}&{83.80}({0.04)}&{96.05}({0.04)}&{99.67}({0.06)}&{84.28}({0.00)}&{1.52}({0.08)}&{46.14}({1.77)}&{2.67}({0.11)}&{38.14}({0.44)}\\
{UniSVM8}&{84.38}({0.04)}&{94.76}({0.05)}&{99.25}({0.10)}&{84.39}({0.00)}&{0.76}({0.07)}&{23.32}({0.48)}&{0.99}({0.05)}&{36.53}({0.44)}\\
{UniSVM9}&{\textbf{85.09}}({0.02)}&{\textbf{97.68}}({0.04)}&{\textbf{99.80}}({0.00)}&{85.31}({0.00)}&{1.06}({0.07)}&{44.67}({2.99)}&{4.29}({0.22)}&{36.55}({0.48)}\\
{UniSVM10}&{84.83}({0.02)}&{95.77}({0.09)}&{99.44}({0.05)}&{\textbf{85.46}}({0.00)}&{0.56}({0.06)}&{21.26}({0.47)}&{0.81}({0.04)}&{35.69}({0.46)}\\
\bottomrule
\end{tabular}}}

\end{table}

From the results in Table \ref{exp:class} and Table \ref{exp:class2}, we conclude the following findings:
\begin{itemize}
  \item UniSVMs with different losses work well using a unified scheme in all cases. They are mostly faster than LibSVM and offer comparable performance in free-noise cases. The training time of LibSVM in Table \ref{exp:class2} is notably longer than its training time in Table \ref{exp:class} because the flipping process increases a large number of support vectors. However, owing to the sparse solution of \eqref{eq:alpha_sparse}, this influence on UniSVMs is quite weak.

  \item Comparing with the training time (including the time to obtain $\PM$ for approximating the kernel matrix $\K$) of UniSVM$_1$ (least squares) with others, it is clear that the proposed UniSVM requires a very low cost after the first iteration, as other UniSVMs always run UniSVM$_1$ in their first iteration.
  \item All the UniSVMs with nonconvex loss are working as efficiently as those with convex loss. Particularly, the UniSVMs with nonconvex losses maintain high performance on the contaminated data sets. The new proposed loss \eqref{eq:los4} with two more parameters always achieves the highest performance.
\end{itemize}

The second set of experiments examined the performance of the UniSVM for solving regression tasks with convex and nonconvex losses. The experimental results are listed in Table \ref{exp:regression}. The chosen losses for UniSVM$_1$ to UniSVM$_{6}$ are listed as follows:\\
\vspace{.4cm}
\resizebox{\textwidth}{!}{
\begin{tabular}{ll}
 UniSVM$_1$: least squares loss, &UniSVM$_2$: smoothed $\varepsilon$-insensitive loss \eqref{eq:smooth_insensitive loss} ($p=100$),\\
 UniSVM$_3$: Hubber loss ($\delta=0.1$ ), &UniSVM$_4$: smoothed absolute loss ($p=100$),
 \\
 UniSVM$_5$: truncated least squares ($a=2$), &UniSVM$_6$: truncated Hubber loss ($\delta=0.1,a=2$).\\
\end{tabular}}

\begin{table}[htp]
  \centering
  \caption{\textbf{Regression task}--Test RMSE (root-mean-square-error) and the training time of the related algorithms on the benchmark data sets. All results are averaged over ten trials with the standard deviations in brackets; The first four lines are based on convex losses and the rest are based on the truncated nonconvex losses.}\label{exp:regression}
\resizebox{\textwidth}{!}
{
\setlength{\tabcolsep}{3mm}
{
\begin{tabular}{rcccccc}
\toprule
&\multicolumn{3}{c}{Test RMSE } & \multicolumn{3}{c}{Training time (CPU seconds)}\\
\cmidrule(r){2-4} \cmidrule(r){5-7}
Algorithm&Cadata&3D-Spatial&Slice&Cadata&3D-Spatial&Slice\\
\cmidrule(r){1-7}
{LibSVM~~~}&{0.314}({0.000)}&{0.464}({0.000)}&---&{3.38}({0.04)}&{4165}({2166)}&$>3hr$\\
{UniSVM$_{1}$}&{0.314}({0.000)}&{0.455}({0.000)}&{\textbf{6.725}}({0.101)}&{1.06}({0.06})&{\textbf{96.3}}({5.0)}&{\textbf{25.40}}({0.08)}\\
{UniSVM$_{2}$}&{\textbf{0.307}}({0.000)}&{0.459}({0.000)}&{6.753}({0.100)}&{1.38}({0.07)}&{118.9}({4.2)}&{44.75}({1.05)}\\
{UniSVM$_{3}$}&{0.310}({0.000)}&{0.463}({0.000)}&{6.870}({0.103)}&{1.31}({0.09)}&{112.8}({3.8)}&{60.82}({1.70)}\\
{UniSVM$_{4}$}&{0.308}({0.000)}&{0.464}({0.000)}&{6.765}({0.100)}&{1.44}({0.08)}&{123.3}({4.3)}&{50.89}({1.25)}\\
\cmidrule(r){1-7}
{UniSVM$_{5}$}&{0.315}({0.000)}&{\textbf{0.454}}({0.000)}&{6.868}({0.116)}&{1.10}({0.06)}&{99.6}({4.1)}&{83.75}({13.30)}\\
{UniSVM$_{6}$}&{0.312}({0.000)}&{0.465}({0.000)}&{6.775}({0.105)}&{1.32}({0.07)}&{121.5}({4.9)}&{75.72}({4.45)}\\
\bottomrule
\end{tabular}
}}
\end{table}

From the results in Table \ref{exp:regression}, it is again observed that UniSVMs with different losses work well for a unified scheme. All of them are more efficient than LibSVM with comparable performance. For example, LibSVM costs very long training time on the second 3D-Spatial task because of excessive training samples, and LibSVM cannot finish the task on the last Slice data set, possibly because of excessive support vectors. In two cases, all UniSVMs function well and exhibit comparable performance, which is primarily attributed to the efficient low-rank approximation of the kernel matrix. It is also noted that the UniSVMs with nonconvex losses function as efficiently as those with convex loss.

In the third set of experiments, we challenge UniSVM with two classification tasks on the very large data sets (up to millions of samples) on the same computer. The selected data sets are:
\begin{itemize}
  \item \texttt{Covtype}: a binary class problem with
581,012 samples, where each example has 54 features. We randomly split it into 381,012 training
samples and 200,000 test samples. The parameters used are $\gamma=2^{-2}$ and $\lambda=10^{-8}$.
  \item \texttt{Checkerboard3M}: based on the noise-free version of the 2-dimensional Checkerboard data set ($4\times4$-grid XOR problem), which was widely used to show the effectiveness of nonlinear kernel methods. The data set was sampled by uniformly discretizing the regions $[0,1]\times[0,1]$ to $2000^2 = 4000000$ points and labeling two classes by the $4\times4$-grid XOR problem, and was then split randomly into 3,000,000 training samples and 1,000,000 test samples. The parameters used are $\gamma=2^{4}$ and $\lambda=10^{-7}$.
\end{itemize}
Those data sets are also used in \cite{Zhou2016}. Because of the limited memory of our computer, the kernel matrix on \texttt{Covtype} is approximated as $\PM\PM^\top$ with $\PM\in\Real^{m\times 1000}$, and the kernel matrix on \texttt{Checkerboard3M} is approximated as $\PM\PM^\top$ with $\PM\in\Real^{m\times 300}$, where $m$ is the training size.
The experimental results are given in Table \ref{exp:class3}, where LibSVM cannot accomplish the tasks because of its long training time. The losses used in the algorithms are the same as those in Table \ref{exp:class}.

\begin{table}[htp]
  \centering
  \caption{\textbf{Classification III}--Test accuracies and training times of the related algorithms on two very large data sets, \texttt{Covtype} and \texttt{Checkerboard3M}, where all results are averaged over \emph{five} trials with the standard deviations in brackets. The first three lines are based on convex losses and the others are based on nonconvex losses.}\label{exp:class3}
\resizebox{\textwidth}{!}
{
\setlength{\tabcolsep}{3mm}
{
\begin{tabular}{rcccc}
\toprule
&\multicolumn{2}{c}{Test accuracy (\%)}&\multicolumn{2}{c}{Training time (CPU seconds)}\\
\cmidrule(r){2-3}\cmidrule(r){4-5}
Algorithm&Covtype &Checkerboard3M&Covtype &Checkerboard3M\\
\cmidrule(r){1-5}
{UniSVM$_{1}$}&{81.11}({0.02)}&{98.04}({0.08)}&{\textbf{183.68}}({11.80)}&\textbf{{37.94}}({2.68)}\\
{UniSVM$_{2}$}&{80.80}({0.03)}&{98.05}({0.18)}&{205.92}({12.77)}&{77.28}({2.73)}\\
{UniSVM$_{3}$}&{81.14}({0.02)}&{98.07}({0.08)}&{188.40}({12.17)}&{40.72}({2.67)}\\
\cmidrule(r){1-5}
{UniSVM$_{4}$}&{83.15}({0.12)}&{99.94}({0.01)}&{540.00}({84.54)}&{634.54}({45.18)}\\
{UniSVM$_{5}$}&{81.46}({0.04)}&{97.99}({0.07)}&{224.34}({15.00)}&{42.53}({2.87)}\\
{UniSVM$_{6}$}&{83.25}({0.14)}&{99.94}({0.01)}&{449.73}({42.14)}&{574.43}({4.22)}\\
{UniSVM$_{7}$}&{82.90}({0.10)}&{99.83}({0.03)}&{405.50}({11.54)}&{545.35}({3.81)}\\
{UniSVM$_{8}$}&{82.19}({0.09)}&{99.90}({0.01)}&{282.55}({16.65)}&{580.34}({3.76)}\\
{UniSVM$_{9}$}&{\textbf{83.40}}({0.05)}&{\textbf{99.95}}({0.01)}&{409.71}({44.09)}&{693.48}({4.30)}\\
{UniSVM$_{10}$}&{81.89}({0.03)}&{99.94}({0.02)}&{269.06}({10.15)}&{777.14}({8.63)}\\
\bottomrule
\end{tabular}}}
\end{table}

From the results in Table \ref{exp:class3}, we observe that the UniSVM works well on very large data sets. We also reach conclusions which are consistent with the results in Tables \ref{exp:class} and \ref{exp:class2}. For example, UniSVMs with different losses work well by a unified scheme and offer comparable performance, and all the UniSVMs with nonconvex loss function as efficiently as those with convex loss. Particularly, the UniSVMs with nonconvex losses maintain high performance because many contaminated samples may exist in very large training cases.

\section{Conclusion and future work}\label{sec:con}
In this work, we first define a kind of LS-DC loss with an effective DC decomposition. Based on the DCA procedure, we then propose a unified algorithm (UniSVM) for training SVM models with different losses for classification problems and for the regression problems.
Particularly, for training robust SVM models with nonconvex losses, UniSVM has a dominant advantage over all the existing algorithms because it always has a closed-form solution per iteration, while the existing ones must solve a constraint programming per iteration. Furthermore, UniSVM can solve the large-scale nonlinear problems with efficiency after the kernel matrix has the low-rank matrix approximation.

Several experimental results verify the efficacy and feasibility of the proposed algorithm. The most prominent advantage of the proposed algorithm is that it can be easily grasped by users or researchers since its core code in Matlab is less than 10 lines (See Appendix \ref{appendix:smallcase}).

In this work, we mainly discussed the methods to deal with the (convex or nonconvex) loss of the regularized loss minimization \citep{Shalev-Shwartz2014} by DCA to enhance the sparseness of the samples or robustness of the learner. However, there are also some works which handle the nonconvex regularizer part of the regularized loss minimization by DCA, which can strengthen the sparseness of the features and serve as a highly efficient tool for feature selection. For example, in \citet{Neumann2004,Le2008,Le2009, Ong2013}, some smooth approximations of the nonconvex ``$\ell_0$ norm'' are decomposed as DC forms, then DCA is used to perform feature selection and many satisfactory results are produced. We will intensively study whether or not our new LS-DC decomposition can improve those kinds of learning problems.

\begin{acknowledgements}{We would like to acknowledge support for this project from the National Natural Science Foundation of China under Grant No. 61772020. We also thank the anonymous reviewers for their useful comments that greatly improved the presentation.}
\end{acknowledgements}

\section*{Conflict of interest}

The authors declare that they have no conflict of interest.

\appendix
\section*{Appendix}
\section{The proof of Propositions}\label{appendix:proof}
\subsection{The proof of Propositions 1}
\begin{proof}
We illustrate them one by one.
\begin{enumerate}[(a)]
\item It is clear.
\item It is because $u^2-\min\{u^2, a\}= (u^2 - a)_+$ is a convex function.
\item It is because $Au^2-u_+^2$ with $A\geq1$ is a convex function.
\item It is because $Au^2-\min\{u_+^2,a\} = Au^2-u_+^2 + (u_+^2-a)_+$ with $A\geq1$ is convex.
\item First we show that the hinge loss $\ell(y,t) = (1-yt)_+$ is \textbf{not} an LS-DC loss. If let $g(u)=Au^2-u_+$, we have $g_-'(0)=0>g'_+(0)=-1$. Hence by Theorem 24.1 in \cite{Rockafellar1972}, we conclude that $g(u)$ is not convex for all $A$ ($0<A<+\infty$).

    Noticed that  $(1-yt)_+ = \lim_{p\rightarrow+\infty}\frac1p\log(1+\exp(p(1-yt)))$. Let $\psi(u) = \frac1p\log(1+\exp(pu))$, and we have $\psi''(u) = \frac{p\exp(pu)}{(1+\exp(pu))^2}\leq \frac{p}{4}$. By Theorem \ref{th:th1}, we know that $\ell_p(y,t)=\frac1p\log(1+\exp(p(1-yt)))$ is an LS-DC loss with $A\geq p/8$. In experiments, letting $1\leq p\leq 100$, $\ell_p(y,t)$ is a  good approximation of the hinge loss.
\item The reason that the ramp loss is \textbf{not} an LS-DC loss is the same as that of the hinge loss. It's two smoothed approximations \eqref{eq:ramp_s} and \eqref{eq:ramp_sp} are LS-DC loss. The proof of the first is similar as that of the squared hinge loss and the proof of second is similar as that of the approximation of hinge loss.
\item Let $g(u) = Au^2-a\left(1-\exp(-\tfrac1 au_+^2)\right)$. Then $g(u)$ is a convex function because $g'(u)=2Au-2u_+\exp(-\tfrac1au_+^2))$ is monotonically increasing if $A\geq1$.
\item Let $\psi(u)=a(1-e^{-\tfrac1b u^c_+})$. If $c=2$, set $A\geq \frac a b=\frac12 M(a,b,2)$ and let $g(u) = A u^2-\psi(u)$. Hence $g(u)$ is convex because $g'(u)=2A u- \frac{2a}b u_+ e^{-\frac1b u^2_+}$ is monotonically increasing.

    For $c>2$,  $\psi(u)$ is second-order derivable, according to Theorem \ref{th:th1} we only need to obtain the upper bound of $\psi''(u)$. If $u\leq 0$, we have $\psi''(u)=0$. If $u>0$, then $\psi''(u)=\tfrac{ac}{b} \left((c-1)u^{c-2}-\tfrac{c}{b} u^{2c-2}\right) e^{-\frac1b u^c}$ and
      $$\psi'''(u)=\tfrac{ac}{b} u^{c-3}e^{-\frac1b u^c}\left(\tfrac{c^2}{b^2}u^{2c}-\tfrac{3c(c-1)}{b} u^{c}+(c-1)(c-2)\right).$$
    Letting $\psi'''(u)=0$, we get the roots $u^*_1$ and $u^*_2$ ($0<u^*_1<u^*_2$), where
    $$u^*_{1} = (b\cdot h(c))^{\frac1c}$$
    with $h(c)=(3(c-1) - \sqrt{5c^2-6c+1})/(2c)$, which is a local maximum of $\psi''(u)$.
    Noting that $\lim_{u\rightarrow0} \psi''(u) = \lim_{u\rightarrow \infty}\psi''(u) =0$, we have that the global maximum of $\psi''(u)$ reaches at $u^*_1$. Putting $u^*_1$ in $\psi''(u)$, we prove that $\psi''(u)\leq \psi''(u^*_1)=:M(a,b,c)$ for any $u$, where $M(a,b,c)=\tfrac{ac}{b^{2/c}}\left((c-1)(h(c))^{1-2/c}-c (h(c))^{2-2/c}\right)e^{-h(c)}$.

    For example, $M(2,2,2) = 2$, $M(2,2,4) \approx 4.5707<5$, $M(2,3,4) \approx 3.7319<4$. Thus, the parameter $A= \frac12 M(a,b,c)$ is not very large.
\end{enumerate}
\end{proof}

\subsection{The proof of Propositions 2}
\begin{proof}
We illustrate them one by one.
\begin{enumerate}[(1)]
  \item It is clear.
  \item The $\varepsilon$-insensitive loss $\ell_\varepsilon(y,t):= (|y-t|-\varepsilon)_+$ is \textbf{not} an LS-DC loss. The reason is similar as that of the hinge loss in item (\ref{item:6}). However, its smoothed approximation \eqref{eq:smooth_insensitive loss} is LS-DC loss with $A\geq p/4$. Actually, let $\psi(u) = \frac{1}{p}\log(1+\exp(-p(u+\varepsilon))) + \frac{1}{p}\log(1+\exp(p(u-\varepsilon)))$. We have $$\psi''(u) = \frac{p\exp(-p(u+\varepsilon))}{(1+\exp(-p(u+\varepsilon)))^2} + \frac{p\exp(p(u-\varepsilon))}{(1+\exp(p(u-\varepsilon)))^2}\leq \frac p 2.$$

    \item The absolute loss $\ell(y,t) = |y-t|$ is also \textbf{not} an LS-DC loss. Clearly, the Hubber loss $\ell_\delta(y,t)$ which approximates the absolute loss, is an LS-DC loss with $A\geq 1/(2\delta)$; Setting $\varepsilon=0$ in \eqref{eq:smooth_insensitive loss} we obtain another smoothed absolute loss, which is an LS-DC loss with $A\geq p/4$.
    \item It is clear.
\end{enumerate}
\end{proof}

\section{The lists of some related losses and their subdifferentials}\label{appendix_a}
The most related losses and their subdifferentials for updating $\vgamma^k$ by \eqref{eq:gamma_update} are listed in Table \ref{tab:classification_loss}. The LS-DC parameters of the LS-DC losses are also given in last column. 
In experiments, we always use the lower-bound of the parameter. 

\begin{table}[H]
\newcommand{\tabincell}[2]{\begin{tabular}{@{}#1@{}}#2\end{tabular}}
\renewcommand\arraystretch{1.05}
\caption{The list of the losses and their subdifferentials (partly).}
  \label{tab:classification_loss}
\resizebox{1.0\textwidth}{!}
{
\setlength{\tabcolsep}{1mm}{
\begin{tabular}{m{2cm}|l|l|c}

\toprule
&\multicolumn{3}{c}{\textbf{Classification losses}:  $\ell(y, t)=\psi(1-yt)$ and $\frac\partial{\partial t}\ell\left(y, t\right) = -y\partial\psi(1-yt)$.}\\
\cmidrule(r){2-4}
Loss name  & $\psi(u)$ &$\partial\psi(u)$ &\tabincell{c}{$A$}\\
\cmidrule(r){1-4}

\rowcolor[gray]{0.9}%
 Least squares loss
&$\psi^{(1)}(u):= u^2$
&$\partial\psi(u):=2u$& $\ge1$\\

Truncated Least squares loss
&$\psi^{(2)}_a(u):=\min\{u^2,a\}$
&$\partial\psi_a(u):=\left\{\begin{array}{ll}
                              2u, & |u|<\sqrt{a}, \\
                              0, & |u|\geq\sqrt{a},
                            \end{array}\right.$\\
 \rowcolor[gray]{0.9}%

Squared hinge loss
&$\psi^{(3)}(u):=u_+^2$
&$\partial\psi(u):=2u_+$&$\ge1$\\

Truncated squared hinge loss
&$\psi^{(4)}_{a}(u):=\min\{u_+^2, a\}$
&$\partial\psi_{a}(u):=\left\{\begin{array}{ll}
                              2u, & 0<u<\sqrt{a}, \\
                              0, & others,
                            \end{array}\right.$\\
 \rowcolor[gray]{0.9}

Hinge loss
&\multicolumn{3}{l}{$\psi^{(5)}(u):=u_+$,
 NOT LS-DC loss, smoothed by $\psi^{(6)}$.}\\

Smooth Hinge loss
&$\psi^{(6)}_{p}(u) :
= u_+ + \frac{\log\left(1+e^{-p|u|}\right)}p$
&$\partial\psi_{p}(u) :
= \frac{\min\{1,e^{pu}\}}{(1+e^{-p|u|)}}$
&$\ge\frac p 8$\\
\rowcolor[gray]{0.9}
Ramp loss
&\multicolumn{3}{l}{$\psi^{(7)}_{a}(u):=\min\{u_+,a\}$, NOT LS-DC loss, smoothed by $\psi^{(8)}$ and $\psi^{(9)}$.}\\

Smoothed ramp loss 1
&$\psi^{(8)}_{a}(u):=\left\{\begin{array}{ll}
                              \frac2a u_+^2, & u\leq\frac {a} 2, \\
                              a-\frac2a (a-u)_+^2, & u>\frac {a} 2,
                            \end{array}\right.
$
&$\partial\psi_{a}(u):=\left\{\begin{array}{ll}
                              \frac4a u_+, & u\leq\frac {a} 2, \\
                              \frac4a (a-u)_+, & u>\frac {a} 2,
                            \end{array}\right.
$\\
\rowcolor[gray]{0.9}
Smoothed ramp loss 2
&$\psi^{(9)}_{(a,p)}(u):=\tfrac1p\log\left(\frac{1+e^{pu}}{1+e^{p(u-a)}}\right)$
&$\partial\psi_{(a,p)}(u):=\frac{e^{-p(u-a)}-e^{-pu}}{(1+e^{-p(u-a)})(1+e^{-pu})} $
&$\ge\frac p 8$\\

smoothed nonconvex loss\eqref{eq:los4}
&$\psi^{(10)}_{(a,b,c)}(u):=a\left(1-e^{-\tfrac1b u_+^c}\right)$
&$\partial\psi_{(a,b,c)}(u):=\frac{ac}{b} u_+^{c-1} e^{-\frac1b u_+^c}$ & \tabincell{c}{$\ge\frac12M(a,b,c)$\\ See \eqref{eq:M_abc}}\\

\cmidrule(r){1-4}
&\multicolumn{3}{c}{\textbf{Regression losses}: $\ell(y, t)=\tilde\psi(y-t)$ and $\frac\partial{\partial t}\ell\left(y, t\right) = -\partial\tilde\psi(y-t)$.}\\
\cmidrule(r){2-4}
\rowcolor[gray]{0.9}
Least squares loss
&$\tilde\psi^{(1)}(u):=u^2$
&$\partial\tilde\psi(u):=2u$& $\ge1$\\

Truncated least squares loss
&$\tilde\psi^{(2)}_{a}(u): = \min\{u^2,a\}$
&$\partial\tilde\psi_{a}(u): = \left\{\begin{array}{ll}
                              2u, & |u|<\sqrt{a}, \\
                              0, & |u|\geq\sqrt{a},
                            \end{array}\right.$ &$\ge1$\\

\rowcolor[gray]{0.9}
$\varepsilon$-insensitive loss
&\multicolumn{3}{l}{ $\tilde\psi^{(3)}_\varepsilon(u):=(|u|-\varepsilon)_+$, NOT LS-DC loss, smoothed by $\tilde\psi^{(4)}$.}\\

smoothed $\varepsilon$-insensitive loss
&$\tilde\psi^{(4)}_{(p,\varepsilon)}(u):=\frac{\log((1+e^{-p(u+\varepsilon)})(1+e^{p(u-\varepsilon)}))}p$&
 $\partial\tilde\psi_{(p,\varepsilon)}(u) :=\frac{1}{1+e^{p(u-\varepsilon)}}- \frac{1}{1+e^{-p(\varepsilon+u)}}$
  &$\ge\frac p 4$\\
\rowcolor[gray]{0.9}
Absolute loss
&\multicolumn{3}{l}{ $\tilde\psi^{(5)}(u):=|u|$, NOT LS-DC loss, smoothed by $\tilde\psi^{(6)}$ and $\tilde\psi^{(7)}$.}\\

Huber loss
& $\tilde\psi^{(6)}_\delta(u):=\left\{\begin{array}{ll}
                    \frac1{2\delta} u^2, & |u|<\delta, \\
                    |u|-\frac{\delta}{2},& |u|\geq\delta,
                                            \end{array}
\right.$
&$\partial\tilde\psi_\delta(u):=\left\{\begin{array}{ll}
                    \frac1{2\delta} u, & |u|<\delta, \\
                    {\rm sgn}(u),& |u|\geq\delta,
                                            \end{array}
\right.$
&$\ge\frac1{2\delta}$\\
\rowcolor[gray]{0.9}
Smoothed Absolute loss
&$\tilde\psi^{(7)}_p(u):=\frac1p\log((1+e^{-pu})(1+e^{pu}))$
&$\partial\tilde\psi_{p}(u) :=\frac{\min\{1, e^{pu}\}-\min\{1, e^{-pu}\}}{1+e^{-p|u|}}$
&$\ge\frac p 4$\\

Truncated Hubber loss
& $\tilde\psi^{(8)}_{(\delta,a)}(u):=\min\{\tilde\psi^{(6)}_\delta(u), a\}$
&$\partial\tilde\psi_\delta(u):=\left\{\begin{array}{ll}
                    \frac1{2\delta} u, & |u|<\delta, \\
                    {\rm sgn}(u),& \delta\leq|u|\leq a,\\
                    0, &   |u|> a.
                    \end{array}\right.$
&$\ge\frac 1{2\delta}$\\
\bottomrule
\end{tabular}
}}
\end{table}

\section{Matlab code for UniSVM}\label{appendix:smallcase}

Matlab code for solving UniSVM with the full kernel matrix available in Subsection \ref{sec:sub_smallcase} is listed as follows, and see the notes for other cases.
The demo codes can also be found on site \texttt{https://github.com/stayones/code-UNiSVM}.
\begin{verbatim}
0: function [alpha] = UniSVM_small(K, y, lambda, A, dloss, eps0)
1: %K-kernel matrix; y-targets; lambda-regularizer;
   %A-parameter of the LS-DC loss; dloss-the derivative function of the LS-DC loss.
2: m = length(y);  v_old = zeros(m,1);
3: Q = inv(K + lambda * m / A * eye(m));alpha = Q*y; %This is the LS-SVM solution.
4: while 1
5:    Ka =  K * alpha;
6:    v= - y .* dloss(1-y .* Ka); %for CLASSIFICATION task;
      % v= - dloss(y - Ka);       %for REGRESSION task;
7:    if norm(v_old - v) < eps0, break; end
8:    alpha = Q * (Ka - v *(0.5/A)) ;  v_old = v;%
9: end
   return

Note:
1) With squared hinge loss, dloss(u)=2*max(u,0);
   With truncated squared hinge loss, dloss(u)=2*max(u,0).*(u<=sqrt(a));
   With truncated least squares loss, dloss(u)=2*(u).*(abs(u)<=sqrt(a));
   With other losses, dloss(u) given in the Table 5 in Appendix B.
2) For large training problem, the input K is taken place as P and B with K=P*P',
   then make the following revisions:
   Line 3 --> Q=inv((lambda*m/A*eye(length(B)) + P'*P)*P(B,:)');alpha = Q*(P'*y);
   Line 5 --> Ka = P*(P(B,:)'*alpha);
   Line 8 --> alpha = Q*(P'*(Ka - v *(0.5/A))); v_old = v.
\end{verbatim}

\end{document}